\documentclass[a4paper,11pt]{article} 
\usepackage{geometry}
\usepackage{caption}
\usepackage{float}
\usepackage{subcaption}
\usepackage{graphicx}
\usepackage{adjustbox}
\usepackage{microtype}
\usepackage{array}
\newcolumntype{H}{>{\setbox0=\hbox\bgroup}c<{\egroup}@{}}

\usepackage{tabularx}
\usepackage{fancyvrb}

\usepackage{booktabs}
\usepackage{amsmath,amsthm,thmtools,amssymb}
\usepackage{blkarray}
\usepackage{nicefrac}

\newcommand{\mneg}{\bar}

\newtheorem{example}{Example}

\declaretheoremstyle[
  headfont=\normalfont\scshape,
  numbered=unless unique,
  bodyfont=\normalfont,
  spaceabove=1em plus 0.75em minus 0.25em,
  prefoothook=\hfill\ensuremath{\blacksquare},%
  spacebelow=1em plus 0.75em minus 0.25em,
]{exmpstyle}

\declaretheorem[
  style=exmpstyle,
  title=Example,
  refname={example,examples},
  Refname={Example,Examples}
]{exmp}
\renewenvironment{example}{\begin{exmp}}{\end{exmp}}

\usepackage{xspace}
\newcommand{\SAT}{\texttt{SAT}\xspace}
\newcommand{\MC}{\texttt{MC}\xspace}
\newcommand{\WMC}{\texttt{WMC}\xspace}
\newcommand{\PMC}{\texttt{PMC}\xspace}
\newcommand{\PWMC}{\texttt{PWMC}\xspace}
\newcommand{\ta}[1]{2^{#1}}
\newcommand{\eqdef}{\ensuremath{\,\mathrel{\mathop:}=}}
\newcommand{\SB}{\{}%
\newcommand{\SM}{\mid}%
\newcommand{\SE}{\}}%

\newcommand{\Card}[1]{\left|#1\right|}

\DeclareMathOperator{\mc}{mc}
\DeclareMathOperator{\wmc}{wmc}
\DeclareMathOperator{\pmc}{pmc}
\DeclareMathOperator{\pwmc}{pwmc}
\DeclareMathOperator{\var}{var}
\DeclareMathOperator{\lits}{lits}
\DeclareMathOperator{\Mod}{Mod}
\DeclareMathOperator{\PMod}{PMod}

\newcommand{\problemFont}[1]{\textsc{#1}}
\newlength\problemlength
\newcommand\dproblem[3]{%
\begin{center}
\fbox{%
\begin{minipage}{.98	\linewidth}%
\begin{list}{}{\labelwidth\problemlength \labelsep.7em \rightmargin1.5em
\leftmargin\problemlength \advance\leftmargin by3em%
\parsep0ex \itemsep.2ex plus.1ex}
\item[{\sl Problem:\hfill}] {\problemFont{#1}}
\item[{\sl Input:  \hfill}] #2
\item[{\sl Task: \hfill}] #3
\end{list}
\end{minipage}
}
\end{center}
}

\usepackage{xcolor}
\newcommand{\solver}[1]{\texttt{#1}}
\newcommand{\ops}[1]{{``\mbox{#1}''}}

\DeclareMathOperator{\cntc}{\#\cdot}%

\newcommand{\complexityclass}[1]{\ensuremath{\mathsf{#1}}}
\newcommand{\NP}{\ensuremath{\complexityclass{NP}}}

\newcommand{\cP}{\ensuremath{\cntc\complexityclass{P}}}
\newcommand{\sharpP}{\ensuremath{\#\complexityclass{P}}}
\newcommand{\cNP}{\ensuremath{\cntc\NP}}

\usepackage{hyperref}

\DeclareMathOperator{\dom}{\mathsf{dom}}%
\newcommand{\fsym}[1]{\texttt{#1}}

\newcommand{\err}{\ensuremath{\epsilon_\text{max}}}
\newcommand{\wrg}{\ensuremath{w_\text{max}}}
\newcommand{\errc}{\ensuremath{\epsilon}}
\newcommand{\RLPD}{\text{lpc}}%
\newcommand{\unk}{\ensuremath{-1}\xspace}%
\newcommand{\cnt}{\ensuremath{c}}%

\newcommand{\cm}[0]{\checkmark}

\newcommand{\sources}{\nolinkurl{Sources}}
\newcommand{\binary}{\nolinkurl{Binary}}

\newcommand{\bddm}{bdd\_{}minisat\_{}all}
\newcommand{\exact}{E}
\newcommand{\arb}{E$^*$}
\newcommand{\app}{A}
\newcommand{\pf}{$^\dagger$}
\newcommand{\arch}{$^\ddagger$}
\usepackage[affil-it]{authblk}
\usepackage[misc]{ifsym}%
\newcommand{\email}[1]{\texttt{#1}}

\title{The Model Counting Competitions 2021--2023\thanks{See: \url{https://mccompetition.org/}, \url{https://zenodo.org/communities/modelcounting},\\ \Letter: \email{johannes.fichte@liu.se}, \email{hecher@cril.fr}}} 
\author{Johannes K. Fichte}
\affil{Link\"oping University}
  
\author{Markus Hecher}
\affil{Université d'Artois, CNRS, CRIL, France}

\begin{document}
\maketitle

\begin{abstract}
    Modern society is full of computational challenges that rely on
    probabilistic reasoning, statistics, and combinatorics.
    Interestingly, many of these questions can be formulated by
    encoding them into propositional formulas and then asking for its
    number of models.
    With a growing interest in practical problem-solving for tasks
    that involve model counting, the community established the Model
    Counting (MC) Competition in fall of 2019 with its first iteration
    in 2020.
    The competition aims at advancing applications, identifying
    challenging benchmarks, fostering new solver development, and
    enhancing existing solvers for model counting problems and their
    variants.
    The first iteration, brought together various researchers,
    identified challenges, and inspired numerous new applications.

    In this paper, we present a comprehensive overview of the
    2021--2023 iterations of the Model Counting Competition. We detail
    its execution and outcomes.
    The competition comprised four tracks, each focusing on a
    different variant of the model counting problem.
    The first track centered on the model counting problem (MC), which
    seeks the count of models for a given propositional formula.
    The second track challenged developers to submit programs capable
    of solving the weighted model counting problem (WMC).
    The third track was dedicated to projected model counting
    (PMC).
    Finally, we initiated a track that combined projected and weighted
    model counting (PWMC).
    The competition continued with a high level of participation, with
    seven to nine solvers submitted in various different version and
    based on quite diverging techniques.
    
  \end{abstract}

\tableofcontents

\clearpage
\section{Introduction}

\emph{Propositional model counting}, also known as \emph{number SAT}
or \#SAT, asks to output the number of models of a propositional
formula~\cite{BiereEtAl21}. A \emph{model} is also known as total
assignment to the variables of the formula that satisfies the formula.
The \#SAT problem is canonical for the complexity
class~\sharpP~\cite{Valiant79b,Roth96,BacchusDalmaoPitassi03} and we can
employ counting to solve any problem located on the Polynomial Hierarchy
(PH) following from an immediate consequence of Toda's
Theorem~\cite{Toda91a}.
Interestingly, \#SAT remains computationally very hard on several
classes of propositional formulas, where the SAT problem (deciding
satisfiability only) becomes tractable, such as
2-CNFs~\cite{Valiant79b,DahllofJonssonWahlstrom05a}\footnote{\#SAT remains \#P-hard on 2-CNFs under Turing reductions and we expect that its complexity slightly drops under more strict reductions such as log-space many-one reductions~\cite{Immerman98}, which are both theoretically and practically more reasonable.}.
Despite the computational hardness, the field has seen considerable
advances in recent years and highly efficient solvers emerged, capable
of solving larger problems each year~\cite{ChakrabortyMeelVardi16a,
  OztokDarwiche15a, LagniezLoncaMarquis16a, LagniezMarquis17a,
  SharmaEtAl19a, DudekPhanVardi20a}.
These solvers have immediate use in various domains, such as formal
verification of hardware, software, and security protocols,
combinatorial optimization, or computational mathematics.
Over the last few years, various applications can be witnessed in the
literature~\cite{MeelEtAl17a, Fremont19a,ZhaiChenPiskac20,
  Duenas-OsorioMeelParedes17, ShiShihDarwiche20, BalutaChuaMeel21,
  SundermannThumSchaefer20, LatourBabakiFokkinga22,
  FichteGagglRusovac22, KabirEtAl22}.

Due to the increasing practical interest in counting, the Model
Counting Competition (MCC) was conceived in October 2019.
Since the first iteration in 2020~\cite{FichteHecherHamiti21a}, the
competition was executed as an annual event and results presented at
the SAT conference~\cite{sat_conferences23}.
The annual repetition provides a snapshot of the current
state-of-the-art in practical model counting.
Here, we summarize
the last three iterations of the competition (2021--2023 Model
Counting Competitions) and their outcome.

\paragraph{Goals of the Competition} 
The Model Counting Competition aims at deepening the relationship
between the latest theoretical and practical development on
implementations and their practical applications.
It was inspired by the success of the SAT community with persistent
efforts to improve the performance and robustness of solvers in
competitions~\cite{SimonLeBerreHirsch05a,JarvisaloBerreRoussel12a,sat_competition22}.
In the competition, we challenge the community to improve solvers,
identify benchmarks, 
and
facilitate the exchange of novel ideas.
Beyond the interest group, we aim at making a large audience of
researchers aware of counting techniques and their use, particularly,
in cases where enumerating models is expensive and not required but is
often still a common approach.
While researchers and developers compete, there are no monetary prizes, 
and our primary interest is active participation, collaboration, and
long-term improvements that extend the feasibility of model counting
in practice and spark many new applications.

\paragraph{Competition Tracks and Computational Complexity}
The competitions in 2021--2023 featured four tracks. In 2021, Track~1
dealt with model counting (\MC), Track~2 with weighted model counting
(\WMC), Track~3 with projected model counting (\PMC), and Track~4 with
harder model counting instances. In 2022 and 2023, Track~1 highlighted
model counting (\MC), Track~2 weighted model counting (\WMC), Track~3
projected model counting (\PMC), and Track~4 projected weighted model
counting (\PWMC).
For all problems, we take a propositional formula~$F$ in conjunctive
normal form (CNF) as input. Then, the problems are as follows:
\begin{itemize}
\item[(\MC):] The \emph{model counting problem} asks to output the
  number of models for the given formula~$F$.\\[0.25em]
  \MC is known to be \sharpP~\cite{Valiant79b} and
  \cP-complete~\cite{Roth96a} and expected to be much harder than
  \SAT~\cite{Cook71,Levin73}.
\item[(\WMC):] The \emph{weighted model counting}\footnote{\WMC is
    also known as weighted counting, sum-of-product, partition
    function, or probability of evidence.} asks to output the sum of
  weights over all models for the given formula~$F$.
  Therefore, in addition to %
  $F$, the input contains a function that assigns a weight to each
  literal occurring in the formula~$F$.\footnote{A literal is a
    propositional variable or a 
    negated propositional negation,~e.g., $x$ or $\mneg x$.}
  The \emph{weight} of a model~$M$ is the product over the weights of
  the literals in~$M$.\\[0.25em]
  \WMC is also known to be \cP-complete~\cite{Roth96a}.
\item[(\PMC):] The \emph{projected model counting problem} asks to
  output the number of models for the formula~$F$ after restricting each
  model to a set~$P$ of projection variables.
  Therefore, in addition to $F$, %
  the input contains a set~$P$ of variables, called \emph{projection
    variables} or \emph{show variables}.\\[0.25em]
  \PMC is complete for the class
  $\cNP$~\cite{DurandHermannKolaitis05}.

\item[(\PWMC):] The \emph{projected weighted model counting problem}
  asks to output the sum of weights over all models restricted to a
  set~$P$ of projection variables,~i.e., \PWMC combines \WMC and \PMC.\\[0.25em]
  \PWMC is expected to be complete for the class
  $\cNP$,~c.f.~\cite{Roth96a,DurandHermannKolaitis05}.
\end{itemize}

\paragraph{Contributions}
This article focuses on the 2021--2023 editions of the Model Counting
Competitions. To this end, we provide the following insights:

\begin{enumerate}
\item We illustrate the setup, describe the tracks, and summarize the
  outcome of
  the last three iterations of the competition. %
\item We survey the instance sets, which we received over the last
  years following open calls of benchmarks from different application
  settings, collected from various sources, or generated. Furthermore,
  we establish a curated repository of instances, submitted solvers,
  and detailed results. We archived all artifacts in a publicly
  available open research data repository (Zenodo).
  In addition, we provide a basic environment to rerun the
  competitions on common high-performance computing (HPC)
  environments.
\end{enumerate}
The remainder of the paper is organized as follows:
In Section~\ref{sec:prelimns}, we list preliminaries starting with a
basic background and examples in
Section~\ref{sec:prelimns:background}. We illustrate competition
tracks in Section~\ref{sec:prelimns:tracks}, detail considerations
on accuracy and precision in Section~\ref{sec:prelimns:precs},
summarize the computing infrastructure in
Section~\ref{sec:prelimns:infrastructure}, and state submission
requirements in Section~\ref{sec:prelims:requirements}.
We present the collected instances and the selection process in
Section~\ref{sec:instances}.
Then, we summarize outcome of the three iterations as follows:
Track~1~(MC) in Section~\ref{sec:results:mc},
Track~2~(WMC) in Section~\ref{sec:results:wmc},
Track~3~(PMC) in Section~\ref{sec:results:pmc}, and
Track~4~(PWMC) in Section~\ref{sec:results:pwmc}.
We conclude in Section~\ref{sec:concl}.
In addition, we specify the input and output format, give programs for
instance checking and instance generation in~\ref{sec:format}.

\paragraph{Related Works}
We follow the spirit of works in the community of constraint solving
and mathematical problem solving, where already many competitions and
challenges have been organized such as on
ASP~\cite{DodaroRedlSchuller19a},
CSP~\cite{XCSPComp22,SimonisKatsirelosStreeter09a,TackStuckey23a},
SAT~\cite{sat_competition22},
SMT~\cite{BobotBrombergerHoenicke23a}, %
MaxSAT~\cite{JarvisaloEtAl23}, %
UAI~\cite{DechterEtAl22a}, %
QBF~\cite{PulinaSeidlHeisinger23a}, %
and various problem domains such as DIMACS~\cite{dimacs} %
and PACE~\cite{GrossmannHeuerSchulz22}. %
Many solvers for model counting exists and various solvers where
already known prior to the first edition of the model counting
competition~\cite{FichteHecherHamiti21a}.
The solvers are commonly based on techniques from \SAT solving~\cite{%
  GomesKautzSabharwalSelman08a,%
  SangEtAl04,%
  Thurley06a%
}, knowledge compilation~\cite{LagniezMarquis17a}, or approximate
solving~\cite{ChakrabortyEtAl14a,ChakrabortyMeelVardi16a} by means of
sampling using \SAT solvers, and a few solvers employ dynamic
programming~\cite{DudekPhanVardi20b,DudekPhanVardi20a}.
Later, in Table~\ref{tab:mc-solvers}, we provide an overview on
existing solvers and their features.

\paragraph{Solvers}
Table~\ref{tab:mc-solvers} on page~\pageref{tab:mc-solvers} provides a
brief overview on existing solvers. We list solvers that participated
in one of the competitions and solvers that are known in the
literature. The table contains features, hyperlinks, and scientific
references for the solvers.

\begin{table}[H]
  \centering
   \resizebox{0.85\columnwidth}{!}{%
\begin{tabular}{l|c|c|c|c|c|c|rlc}
  \toprule
  Solver                  & \MC & \WMC & \PMC & \PWMC & Year    & Type   & Reference                                              & Download                                                                                                                                             & License   \\
  \midrule                                                                            
  \solver{ADDMC}          & \cm & \cm  &      &       & 2020    & \exact & \cite{DudekPhanVardi20a}                               & \href{https://github.com/vardigroup/ADDMC}{\sources}                                                                                                 & MIT       \\
  \solver{Alt-DPMC}       & \cm & \cm  &      &       & 2023    & \exact & na                                                     & \href{https://github.com/allrtaken/DPMC}{\sources}                                                                                                   & MIT       \\
  \solver{ApproxMC}       & \cm & \cm  & \cm  &       & 2023\pf & \app   & \cite{ChakrabortyEtAl14a}                              & \href{https://github.com/meelgroup/approxmc}{\sources}                                                                                               & MIT       \\
  \solver{c2d}            & \cm & \cm  & \cm  & \cm   & 2022    & \arb   & \cite{Darwiche04a}                                     & \href{http://reasoning.cs.ucla.edu/c2d/download.php}{\binary}                                                                                        & na        \\
  \solver{D4}             & \cm & \cm  & \cm  & \cm   & 2023    & \exact & \cite{LagniezMarquis17a}                               & \href{https://github.com/crillab/D4}{\sources}                                                                                                       & LGPL3     \\
  \solver{DPMC}           & \cm & \cm  &      &       & 2023    & \exact & \cite{DudekPhanVardi20b}                               & \href{https://github.com/vardigroup/DPMC}{\sources}                                                                                                  & MIT       \\
  \solver{ExactMC}        & \cm & \cm  &      &       & 2023    & \exact & \cite{LaiMeelYap21a}                                   & \href{https://github.com/meelgroup/KCBox}{\sources}                                                                                                  & MIT       \\
  \solver{GANAK}          & \cm & \cm  & \cm  &       & 2023    & \exact & \cite{SharmaEtAl19a}                                   & \href{https://github.com/meelgroup/ganak}{\sources}                                                                                                  & MIT       \\
  \solver{GPMC}           & \cm & \cm  & \cm  & \cm   & 2023    & \arb   & \cite{SuzukiHashimotoSakai15a,SuzukiHashimotoSakai17a} & \href{https://git.trs.css.i.nagoya-u.ac.jp/k-hasimt/GPMC}{\sources}                                                                                  & MIT       \\
  \solver{mcTw}           & \cm &      &      &       & 2020    & \exact & na                                                     & \href{https://github.com/swacisko/mc-2020}{\sources}                                                                                                 & MIT       \\
  \solver{miniC2D}        & \cm & \cm  &      &       & 2020    & \arb   & \cite{OztokDarwiche15a}                                & \href{http://reasoning.cs.ucla.edu/minic2d/}{\binary}                                                                                                & na        \\
  \solver{mtmc}           & \cm & \cm  &      &       & 2022    & \exact & na                                                     & \href{https://zenodo.org/records/4899862}{\sources}                                                                                                                        & GPL2      \\
  \solver{pc2dd}          & \cm &      & \cm  &       & 2021    & \exact & \cite{IsogaiHashimotoSakai21a}                         & \href{https://git.trs.css.i.nagoya-u.ac.jp/t_isogai/cnf2bdd}{\sources}                                                                               & MIT       \\
  \solver{ProCount}       & \cm &      &      &       & 2021    & \exact & \cite{DudekPhanVardi21a}                               & \href{https://github.com/vardigroup/DPMC/commits/procount}{\sources}                                                                                 & MIT       \\
  \solver{sharpSAT-td}    & \cm & \cm  & \cm  &       & 2023    & \arb   & \cite{KorhonenJarvisalo21a,KorhonenJarvisalo23a}       & \href{https://github.com/Laakeri/sharpsat-td}{\sources}                                                                                              & MIT       \\
  \solver{SUMC2}          & \cm &      &      &       & 2021    & \exact & \cite{Spence22}                                                     & \href{https://github.com/ivor-spence/sumc}{\sources}                                                                                                 & GPL3      \\
  \midrule
  \solver{Approxcount}    & \cm &      &      &       & --      & \app   & \cite{WeiSelman05a}                                    & \href{http://www.cs.cornell.edu/~sabhar/software/approxcount/approxcount12.tar.bz2}{\sources}                                                        & na        \\
  \solver{Cachet}         & \cm & \cm  &      &       & --      & \exact & \cite{SangEtAl04}                                      & \href{https://web.archive.org/web/20220121083854/https://www.cs.rochester.edu/u/kautz/Cachet/cachet-wmc-1-21.zip}{\sources\arch}                     & zchaff    \\ 
  \solver{\bddm}          & \cm &      &      &       & --      & \arb   & \cite{TodaSoh15a}                                      & \href{http://www.sd.is.uec.ac.jp/toda/code/cnf2obdd.html}{\sources}                                                                                  & MIT       \\
  \solver{BPCount}        & \cm &      &      &       & --      & \app   & \cite{KrocSabharwalSelman11a}                          & na                                                                                                                                                   & na        \\
  \solver{cnf2eadt}       & \cm &      &      &       & --      & \arb   & \cite{KoricheLagniezMarquisThomas13a}                  & \href{http://www.cril.univ-artois.fr/KC/eadt.html}{\binary}                                                                                          & na        \\
  \solver{countAntom}     & \cm &      &      &       & --      & \exact & \cite{BurchardSchubertBecker15a}                       & \href{https://projects.informatik.uni-freiburg.de/projects/countantom}{\sources}                                                                     & MIT       \\
  \solver{dCountAntom}    & \cm &      &      &       & --      & \exact & \cite{BurchardSchubertBecker16a}                       & \href{}{na}                                                                                                                                          & na        \\
  \solver{dpdb}           & \cm & \cm  &      &       & --      & \arb   & \cite{FichteEtAl20}                                    & \href{https://github.com/hmarkus/dp_on_dbs/tree/master}{\sources}                                                                                    & GPL3      \\
  \solver{DMC}            & \cm &      &      &       & --      & \arb   & \cite{LagniezMarquisSzczepanski18a}                    & \href{http://www.cril.univ-artois.fr/KC/dmc.html}{\binary}                                                                                           & na        \\
  \solver{DSHARP}         & \cm &      &      &       & --      & \arb   & \cite{MuiseEtAl12a}                                    & \href{https://github.com/QuMuLab/dsharp}{\sources}                                                                                                   & MIT       \\
  \solver{gpusat}         & \cm & \cm  &      &       & --      & \exact & \cite{FichteEtAl18c}                                   & \href{https://github.com/daajoe/GPUSAT}{\sources}                                                                                                    & GPL3      \\
  \solver{FocusedFlatSAT} & \cm &      &      &       & --      & \app   & \cite{ErmonGomesSabharwal11a}                          & \href{https://web.archive.org/web/20201014145213/https://cs.stanford.edu/~ermon/code/FocusedFlatSAT.zip}{\sources\arch}                              & na        \\
  \solver{nesthdb}        & \cm &      & \cm  &       & --      & \arb   & \cite{HecherThierWoltran20}                            & \href{https://github.com/hmarkus/dp_on_dbs/tree/nesthdb}{\sources}                                                                                   & GPL3      \\         
  \solver{MBound}         & \cm &      &      &       & --      & \app   & \cite{GomesSabharwalSelman06a}                         & \href{https://web.archive.org/web/20211219191941/http://www.cs.cornell.edu/~sabhar/software/xor-scripts/xor-scripts.zip}{\sources\arch}              & na        \\
  \solver{MiniCount}      & \cm &      &      &       & --      & \app   & \cite{KrocSabharwalSelman11a}                          & na                                                                                                                                                   & na        \\
  \solver{projMC}         & \cm &      & \cm  &       & --      & \exact & \cite{LagniezMarquis19a}                               & \href{http://www.cril.univ-artois.fr/KC/projmc.html}{\binary}                                                                                        & na        \\
  \solver{SampleCount}    & \cm &      &      &       & --      & \app   & \cite{GomesHoffmannSabharwal07a}                       & \href{https://web.archive.org/web/20160830063246/http://www.cs.cornell.edu/~sabhar/software/samplecount/samplecount-1.0-04092007.zip}{\sources\arch} & na        \\
  \solver{sdd}            & \cm & \cm  &      &       & --      & \exact & \cite{Darwiche11a}                                     & \href{http://reasoning.cs.ucla.edu/sdd/}{\binary}                                                                                                    & Apache2.0 \\ %
  \solver{sharpCDCL}      & \cm &      &      &       & --      & \exact & na                                                     & \href{https://github.com/conp-solutions/sharpCDCL}{\sources}                                                                                         & MIT       \\
  \solver{sharpSAT}       & \cm &      &      &       & --      & \arb   & \cite{Thurley06a}                                      & \href{https://github.com/marcthurley/sharpSAT}{\sources}                                                                                             & MIT       \\
  \solver{sharpSAT-td-KC} & \cm & \cm  &      &       & --      & \arb   & \cite{KieselEiter23a}                                  & \href{https://github.com/raki123/sharpsat-td}{\sources}                                                                                              & MIT       \\
  \solver{sts}            & \cm &      &      &       & --      & \app   & \cite{ErmonGomesSelman12a}                             & \href{https://github.com/dorcoh/EntropyApproximator}{\sources}                                                                                       & na        \\ 
  \solver{TensorOrder}    & \cm & \cm  &      &       & --      & \exact & \cite{DudekVardi20}                                    & \href{https://github.com/vardigroup/TensorOrder}{\sources}                                                                                           & MIT       \\
  \solver{XorSample}      & \cm &      &      &       & --      & \app   & \cite{GomesSabharwalSelman06b}                         & na                                                                                                                                                   & na        \\
  \bottomrule
\end{tabular}
}%
\caption{%
  Overview on available implementations for model counting.
  Each row states a solver (Solver), its features, %
  the scientific reference,
  a hyperlink (in the PDF) to download the solver sources or its binaries, and
  the solver's licence.
  Solvers in the upper part of the table participated in at least one of the 
  2020--2023 edition of the model counting competition. Its last participation
  is stated in the Year column.
  Solvers in the lower part are available in the literature.
  A symbol ``\cm'' indicates that the solver
  in the respective row solves the problem. \MC refers to model counting, 
  \WMC to weighted model counting, \PMC to projected model counting, 
  and \PWMC to projected weighted model counting. The problems are defined in Section~\ref{sec:prelimns}.
  The Type column states whether the solver implements an approximate algorithm (\app),
  an exact algorithm with potentially small precision loss during handing of numbers (\exact), or
  an exact algorithm with arbitrary precision number handling (\arb).
  For more details on the topic of precision of solvers, we refer to
  Section~\ref{sec:prelimns:precs}.
  ``$\dagger$'' indicates that the solver participated as part of a portfolio.
  ``$\ddagger$'' indicates that the source code or binary is available via
  the \href{https://archive.org/}{Internet Archive}.
  URLs for sources and binaries link to the most recent version.
  For software heritage and replication purposes, we provide an archived data set
  that contains all original submissions in a public data repository~\cite{CERN13a}.
}
\label{tab:mc-solvers}
\end{table}

\section{Preliminaries and Background}\label{sec:prelimns}
In this section, we briefly provide formal notions from propositional
logic and problem definitions.
For a comprehensive introduction and more detailed information, we
refer to other
sources~\cite{BiereHeuleMaarenWalsh09,KleineBuningLettman99}.  
Furthermore, we describe the competition tracks, explain the
requirements for participating and the ranking criteria, as well as
the computing infrastructure used for executing the competition.

\subsection{Background}\label{sec:prelimns:background}
\paragraph{Basics} For a set~$X$, let $\ta{X}$ be the \emph{power set
  of~$X$}.
The domain~$\mathcal{D}$ of a
function~$f:\mathcal{D} \rightarrow \mathcal{A}$ is given
by~$\dom(f)$. By $f^{-1}: \mathcal{A} \rightarrow \mathcal{D}$ we
denote the inverse
function~$f^{-1}:=\SB f(d) \mapsto d \SM d \in \dom(f) \SE$ of a given
function~$f$, if it exists.  
By $\mathbf{e}$ we mean the Euler number and $\log_{10}$ refers to the
logarithm to the base~$10$.

\paragraph{Satisfiability}
Let $U$ be a universe of propositional variables.
A \emph{literal} is a variable~$x$ or its negation~$\mneg x$. 
We call~$x$ \emph{positive} literal and $\mneg x$ \emph{negative}
literal.
A \emph{clause} is a finite set of literals, interpreted as the
disjunction of these literals.  A propositional \emph{formula}~$F$ in
\emph{conjunctive normal form (CNF)} is a finite set of clauses,
interpreted as the conjunction of its clauses.
We let $\var(F)$ and $\lits(F)$ be the set of the variables and set of
literals, respectively, that occur in~$F$.
An \emph{assignment} is a mapping $\tau:X \rightarrow \{0,1\}$ defined
for a set~$X\subseteq U$ of variables.
The formula~$F$ \emph{under assignment~$\tau$} is the formula~$F_\tau$
obtained from~$F$ by (i)~removing all clauses~$c$ that contain a
literal set to~$1$ by $\tau$ and then (ii)~removing from the remaining
clauses all literals set to~$0$ by $\tau$. An assignment~$\tau$
\emph{satisfies} a given formula~$F$ if $F_\tau=\emptyset$.

\paragraph{Models and Counting}
For a satisfying assignment~$\tau$, we call the set~$M$ of variables
that are assigned to true by~$\tau$ a \emph{model} of~$F$,~i.e.,
$M = \SB x \SM x \in \tau^{-1}(1) \SE$.
We let $\Mod(F)$ be the \emph{set of all models} of a
formula~$F$,~i.e.,
$\Mod(F) = \SB \tau^{-1}(1) \SM F_\tau = \emptyset, \tau \in
\ta{\var(F)} \SE$.
Furthermore, we define the \emph{model count}~$\mc(F)$ as the number
of models of the formula~$F$,~i.e., $\mc(F)\eqdef \Card{\Mod(F)}$.
Then, we define the following counting problem:

\dproblem{Model Counting (\MC)}%
{A propositional formula~$F$.}%
{Output the model count~$\mc(F)$, that is, the number of models of the
  formula~$F$.
}
Weighted model counting generalizes model counting and is useful for
probabilistic inference~\cite{ChaviraDarwiche08}.
We let a \emph{weight function}~$w$ be a function that maps each
literal in $F$ to a real between~$0$ and~$1$,~i.e.,
$w: \lits(F) \rightarrow [0,1]$.
Later, we represent $w$ by a $2 \times n$ matrix for
$n=\Card{\var(F)}$ where the first row represents the positive and the
second the negative literal.
Then, for a model~$M\in \Mod(F)$, the \emph{weight of the model~$M$}
is the product over the weights of its literals, meaning,
$w (M) := \prod_{v\in \var(F) \cap M} w(v) \cdot \prod_{v \in \var(F)
  \setminus M} w(\mneg v)$. %
The \emph{weighted model count} (wmc) of formula is the sum of weights
over all its models,~i.e.,
$\wmc(F,w) \eqdef \sum_{M \in \Mod(F)} w(M)$.
Note that if we set all weights to one, that is,
$w: \lits(F) \rightarrow 1$, we have $\wmc(F,w) = \mc(F)$. Same if we
set all weights to 0.5, that is, $w: \lits(F) \rightarrow 1$, we have
$\wmc(F,w) \cdot 2^{\Card{\var(F)}} = \mc(F)$.
We obtain the following problem:

\dproblem{Weighted Model Counting (\WMC)\footnote{
    The problem is sometimes also called sum-of-products, weighted
    counting, partition function, or probability of evidence.
  }%
}%
{A propositional formula~$F$ and a weight
  function~$w: \lits(F) \rightarrow [0,1]$.}%
{Output the weighted model count~$\wmc(F,w)$.}
Next, we define projected counting, which is interesting when formulas
contain auxiliary variables that are not supposed to contribute to the
overall count but need to be checked for satisfiability.
In other words, solutions, that are the same up to auxiliary
variables, count as just one solution.
Therefore, let $P \subseteq \var(F)$ be a set of variables, called
\emph{projection variables} or \emph{show variables} of the
formula~$F$.
We define the projected models of formula~$F$ under the set~$P$ by
$\PMod(F,P) \eqdef \SB M \cap P \SM M \in \Mod(F) \SE$ and 
define \emph{the projected model count}~$\pmc(F,P)$ of the formula~$F$
under the set~$P$ of projection variables by
$\pmc(F,P) \eqdef\Card{ \PMod(F,P)}$.
This gives raise to the following problem:

\dproblem{Projected Model Counting Problem (\PMC)\footnote{Sometimes
    the problem is referred to as $\#\exists\SAT$ and was originally
    coined under the name \#NSAT, for ``nondeterministic SAT'' by
    Valiant~\cite{Valiant79b}.}}%
{A propositional formula~$F$ and a set~$P\subseteq \var(F)$ of
  \emph{projection variables}.}%
{Output the projected model count~$\pmc(F,P)$.}
Finally, we define projected weighted counting.
We assume $w$ to be a weight function and $P$ a set of projection
variables.
Then, we define the \emph{projected weighted model
  count}~$\pwmc(F,P,w)$ of the formula~$F$ under the weight
function~$w$ and the set~$P$ of projection variables be
$\pwmc(F,P) \eqdef \sum_{N \in \PMod(F,P)}w(N)$.

\dproblem{Projected Weighted Model Counting Problem (\PWMC)}%
{A propositional formula~$F$, a set~$P\subseteq \var(F)$ of
  \emph{projection variables}, and a weight function~$w$.}%
{Output the projected weighted model count~$\pwmc(F,P,w)$.}

\bigskip
\noindent
Example~\ref{ex:problems} illustrates the basic definitions below.

\begin{example}\label{ex:problems}
  Consider the formula $F = (a \vee b) \wedge (c \vee d)$, 
  a set~$P=\{a,d\}$ of show variables, and a weight function~$w$. We
  state~$w$ as matrix where columns indicate the variables (a,b,c,d)
  and first row inside the brackets represents the weight for the
  positive literal ($a$, $b$, $c$, $d$) and the second row the
  negative literal ($\mneg a$, $\mneg b$, $\mneg c$, $\mneg d$) as
  indicated by the literals~$x$ and $\mneg x$ right to the brackets,
  in detail:
  \begin{align*}
    w =
    \begin{blockarray}{ccccc}
      a                        & b                         & c                           & d                            \\
      \begin{block}{[cccc]c}
        0.75                   & 0.3                       & 0.8                         & 0.6 & x                      \\
        0.25                   & 0.7                       & 0.2                         & 0.4 & \mneg x                \\
      \end{block}
    \end{blockarray}.
  \end{align*}
  The formula~$F$ has the following models:
  \begin{align*}
    \Mod(F) = \bigg\{
    \begin{array}{cccc}
      \{a,b,c,d\},             & \{a,b,c, \mneg d\},       & \{\mneg a, b, c, d\},       & \{\mneg a , b, c, \mneg d\}, \\
      \{a,b,\mneg c,d\},       & \{a,\mneg b,c, \mneg d\}, & \{\mneg a, \mneg b, c, d\}, &                              \\
      \{a,\mneg b,c,d\},       &                                                                                        \\
      \{a,\mneg b,\mneg c,d\} &                                                                                        \\
    \end{array}
    \bigg\}.
  \end{align*}
  We obtain the following counts.
  The model count is \[\mc(F)=\Card{\Mod(F)}=9.\]
  The weighted model count is
  \begin{align*}
    & \wmc(F,w) \\
    & =  w(\{a,b,c,d\})  +&   w(\{a,b,c,\mneg d\})  +&  \quad\ldots\quad + & w(\{a,\mneg b, \mneg c, d\})\\
              & =  w(a)\cdot w(b)\cdot w(c)\cdot w(d)  +&   w(a)\cdot w(b)\cdot w(c)\cdot w(d)  +&  \quad\ldots\quad + & w(a)\cdot w(b)\cdot w(\mneg c) \cdot w(d)\\
               & =  0.75\cdot 0.3 \cdot 0.8 \cdot 0.6    + & 0.75 \cdot 0.3 \cdot 0.8 \cdot 0.4  +& \quad\ldots\quad + & 0.75\cdot 0.3 \cdot 0.2\cdot 0.6\\ 
               & =  0.759.
  \end{align*}

\noindent  The projected model count is 
  \begin{align*}
    \pmc(F,P) & = \Big| \PMod(F,P) \Big|\\
              & = \Big| \big\{ \{a,b,c,d\} \cap \{a,b\},  \{a,b,c, \mneg d\}\cap \{a,b\}, \quad\ldots\quad,  \{a,\mneg b,\mneg c,d\} \cap \{a,b\}  \big\} \Big| \\ 
              & = \Big| \{a,b\}, \{\mneg a, b\},  \{a,\mneg b\}, \{\mneg a, \mneg b\}   \Big|\\
              & = 4.
  \end{align*}
  Finally, the projected weighted model count is
  \begin{align*}
    \pwmc(F,P,w) %
                  & = w(\{a,b\}) + w(\{\mneg a, b\}) + w(\{a,\mneg b\}) + w(\{\mneg a, \mneg b\})\\
                  & = w(a) \cdot w(b) + w(\mneg a) \cdot w(b) + w(a) \cdot w(\mneg b) + w(\mneg a) \cdot w(\mneg b)\\
                  & = 0.75 \cdot 0.3 + 0.25 \cdot 0.3 + 0.75 \cdot 0.7 + 0.25 \cdot 0.7\\
                  & = 1.
  \end{align*}
\end{example}

\subsection{Competition Tracks}\label{sec:prelimns:tracks}
The competition encompassed four versions of the model counting
problem, which we evaluated in separate tracks.
All tracks require inputs in conjunctive normal form (CNF) given in a
simple text format, which is compatible with the DIMACS-CNF
format~\cite{TrickChvatalCook93a}.
The format extends the format used in SAT
competitions~\cite{BerreRoussel09a,JarvisaloBerreRoussel12a,sat_competition22}
by introducing statements for weights and projections.
Note that the format used in the 2020
competition~\cite{FichteHecherHamiti21a} differs. In 2021, we changed
the format due to multiple requests and to provide compatibility with
SAT-solving.
For more details on the input format, we refer to~\ref{sec:format}.
The recent three iterations of the Model Counting Competition featured
the following tracks:

\begin{itemize}
\item 2021 Competition:
  \begin{itemize}
  \item Track 1: Model Counting (\MC)
  \item Track 2: Weighted Model Counting (\WMC)
  \item Track 3: Projected Model Counting (\PMC)
  \item Track 4: Model Counting (additional harder instances)   
  \end{itemize}
\item 2022/2023 Competitions:
  \begin{itemize}
  \item Track 1: Model Counting (\MC)
  \item Track 2: Weighted Model Counting (\WMC)
  \item Track 3: Projected Model Counting (\PMC)
  \item Track 4: Projected Weighted Model Counting (\PWMC)
  \end{itemize}
\end{itemize}

\subsection{Accuracy, Precision, and Outputs}\label{sec:prelimns:precs}

When considering solutions in model counting two common issues need to
be addressed (i)~accuracy and (ii)~precision~\cite{iso5725}.
(i) \emph{Accuracy} refers to the quality of the solution in terms of
the degree of closeness to the true value, which is commonly affected
by the solving technique. For example, the solver is an approximate
solver or an anytime solver and does not output the exact solution due
to conceptual constructions. While a limitation in accuracy can be
tolerated in many applications, a user needs to be aware whether the
result is exact, approximate, or a lower/upper bound only.
(ii) %
\emph{Precision} refers to obtaining the same results under unchanged
conditions, which in model counting, is primarily affected by the
precision of arithmetic calculations, since we can easily involve very
large integers or fractionals. 
Fast fixed-precision arithmetic implemented in modern hardware
typically offers between 8 and 256 bits of precision. %
Common datatypes according to the IEEE754 standard~\cite{ieee754} such
as decimal128, binary128, or binary256 allow to represent values up to
$1.6113 \cdot 10^{78913}$ (binary256), but have only up to $36$
significant decimal digits.
While we expected that the maximal representable value is sufficient
for model counting, the limitation to $36$ significant decimal digits
might cause a high precision loss.
Since higher precision yields higher space requirements, we expect a
certain performance loss. Hence, developers can choose between
multiple-precision arithmetic or arbitrary-precision arithmetic, which
are built-in with some programming
languages~\cite{java_biginteger97a,ZadkaRossum01a} or can be employed
by using additional libraries~\cite{GranlundEtAl16a}.
To address these situations, we ask that solvers output the expected
accuracy (approximate, exact, heuristic) and the precision (arbitrary,
single precision, double precision, quadruple precision). 
Furthermore, we require the following information in the output. 
The solver needs to specify the notation in which the count is
outputted (log10, float, prec-sci, integer,
fractional). 
Satisfiability of the instance has to be stated to avoid corner cases
in weighted model counting. 
The type of the count needs to be given (model counting, weighted
model counting, projected model counting, or projected weighted model
counting) to avoid potential errors if the problem is identified
incorrectly in the input or solver requires a dedicated flag to set
the problem.
Since the model count can be very large number or the weighted model
count easily indistinguishable from $0$, we ask to output an
estimate~$c_e$ of the model count~$c$ in $\log_{10}$-notation, that
is, $c_e\eqdef \log_{10}(c)$.
Thereby, we avoid over- or underflows and can quickly estimate the
order of a solution without additional arbitrary-precision
computations, which oftentimes require additional libraries that
require extensive preparations on high performance clusters.
We provide more technical details on the mandatory output format
in~\ref{sec:format:output}.

\subsection{Restrictions, Measure, and Ranking}\label{sec:prelimns:infrastructure}\label{sec:ranking}\label{sec:measure}
After we invited developers to participated in the competition, we
organized an online session to discuss needs and receive
suggestions. We incorporated the feedback and updated the 2020 rules
as follows.

\paragraph{Measuring} 
We mainly compare \emph{number of instances where a solution was
  outputted} and \emph{wallclock time}.
We distinguish between (a)~the solver did not output any count and
(b)~the solver believed that it solved an instance and outputted a
count.
In Case~(a), we ignore the instance. In Case~(b), we evaluate whether
the output is acceptable.
An instance will be marked as accepted if the outputted count is
within a relative margin of error or approximation factor depending on
the track or ranking as given in Table~\ref{tab:accuracy}.
In the 2021 Competition, we enforced accuracy requirements by track.
In contrast, in the 2022 Competition, we introduced multiple rankings
and corresponding accuracy to distinguish between different types of
solvers.
We compute the error and approximation factor as follows.
Let $\cnt_o$ refer to the outputted result by the solver, called
\emph{observed count}, and $\cnt_e$ to the pre-computed result, called
\emph{expected count}.
For the expected count~$\cnt_e$, we use an arbitrary precision
integer, fractional, or \unk to state that we could not obtain a count
in the precomputation phase.
In order to avoid a very high error value for very small or large
numbers close to over- or underflow, we employ the log percentage
change to measure the error~\cite{TornqvistVartiaVartia85a}.  The
\emph{log percentage change} is defined %
by
$\RLPD(\cnt_e,\cnt_o) := 10^{\log_{10}(\cnt_{o}) - \log_{10}(\cnt_{\text{e}})}$.
Then, we measure the \emph{error} in
$\errc(\cnt_e,\cnt_o)\eqdef 1-\RLPD(\cnt_e,\cnt_o)$.
We say that the observed count~$\cnt_o$ by a solver for a given
instance is an \emph{accepted count}, if the observed count~$\cnt_o$
is below the maximum expected error~$\err$, more precisely,
$\errc(\cnt_e,\cnt_o) \leq \err$.
Another way to compute an expected range for a solution are
performance guarantees from approximation
algorithms~\cite{Vazirani01a}.
Therefore, for a given $\alpha$, the observed count~$\cnt_o$ is
required to be within a relative error,~i.e.,
$(1+\alpha)$-approximation. 
We say an observed count~$\cnt_o$ is an \emph{approximately accepted
  count}, if the following condition holds:
$\frac{\cnt_e}{1+\alpha}\leq \cnt_o \leq (1 + \alpha)\cdot \cnt_e$.
Using the definitions of accepted counts, we define the \emph{scoring
  function}, which decides the ranking of the solver within the track
of the competition, as the number of acceptably solved instances
together with the number of solutions for instances where no count
could be precomputed, i.e., $\cnt_e = \unk$.
In 2021, we solely employed accepted counts on all tracks and expected
upper bound~$\err$ for an potential error of the solver according to
Table~\ref{tab:accuracy:2021}.
Upon multiple requests, we updated the ranking as presented in
Table~\ref{tab:accuracy:2022ff} in 2022, to incorporate different
capabilities of solvers.
For exact solvers within Rankings~A and B, we use the notion of an
accepted count.
For approximation solvers within Ranking~C, we employ the notion of an
approximately accepted count.

\paragraph{Restrictions}
Run times larger than \emph{3,600} seconds count as \emph{timeout}. We
restrict \emph{main memory (RAM)} to \emph{32GB}.
Temporary disk space is available for input transformation and
preprocessing during the job execution. However, we advise the
developers to use shared memory instead and clean up temporary files
properly as it might account to disk quota otherwise.

\paragraph{Execution}
During the precomputation, instance classification, and
reproducibility phase, we run multiple solvers on the same node in
parallel. However, we ensure that CPUs and memory cacheline are not
over-committed and results are replicable, meaning, we execute at
most~4 solvers in parallel on a node. %
In the competition phase, we run solvers exclusively, meaning that
solvers run sequentially with exclusive access to an entire node and
no other executions have access to that node during that time.
We carried out all three iterations of the competition on the StarExec
system~\cite{StumpSutcliffeTinelli14} using runsolver~\cite{Roussel11}
to control the execution. 
We take runsolver instead of the more precise BenchExec
tool~\cite{BeyerLoweWendler15,BeyerLoweWendler19}, since BenchExec
causes numerous unexpected runtime abortions and quota issues that we
did not manage to debug in time.

\paragraph{Ranking and Disqualification}
For each track, we select 200 from our collected instances.
We describe the selection process in
Section~\ref{sec:instances:selection}.
We split these instances into two sets of 100 instances, which we call
public and private. We publish the \emph{public} set during the
testing phase of the competition and keep the \emph{private} set
secret until the final evaluation.
During the testing phase, we release information to all developers on
the scoring on the public set, but rank only on the private set.
We define the place of a solver within the competition by the number
of solved instances that have an accepted count.
If multiple solvers score in tie, we assign the same place to all
solvers and leave the subsequent place(s) vacant.
In 2021, we applied a very simple ranking scheme that only
incorporates tracks and accepted solutions. 
In 2022/2023, we updated the ranking to incorporate conceptually
differing features of participating solvers and provide a more
representative overview for potential applications.
If a solver outputs too many counts that are not accepted in one
category, we automatically move the solver into a category that allows
for lower precision and more incorrect answers.
In more detail, we used the following rankings:
\begin{itemize}
\item 2021 Competition: Uniform Ranking. More than 20 solutions
  outside margin result in disqualification.
\item 2022/2023 Competitions:
  \begin{itemize}
  \item Ranking A: Exact (arbitrary precision).\\ %
    Any wrong solution results in removal from the ranking.
  \item Ranking B: Exact (small precision loss).\\ %
    More than 20 solutions outside the expected error margin results
    in removal from the ranking.
  \item Ranking C: Approximate (provide approximation guarantee).\\ %
    More than 20 solutions outside the expected approximation factor
    results in in removal from the ranking.
  \item Ranking D: Heuristic, for example, anytime solving.\\
    Every correct answer counts 1 point, otherwise: 0 points.
  \end{itemize}
\end{itemize}

\begin{table}[t]
  \centering
  \begin{subtable}[h]{0.35\textwidth}
    \centering
    \begin{tabular}{lrr}
      \toprule
      Track & \err & \wrg\\
      \midrule
      Track 1 (\MC)  & 0.1\% & $\leq 20$\\
      Track 2 (\WMC) & 1.0\% & $\leq 20$ \\ 
      Track 3 (\PMC) & 1.0\% & $\leq 20$ \\
      Track 4 (\MC) & 20\% & $\leq 20$\\
      \bottomrule
    \end{tabular}
    \caption{%
      2021 Competition requirements enforced by track regardless of
      the type of the solver.
    }\label{tab:accuracy:2021}
  \end{subtable}
  \hspace{1em}
  \begin{subtable}[h]{0.55\textwidth}
    \centering
    \resizebox{1\textwidth}{!}{%
    \begin{tabular}{lrrr}
      \toprule
       Ranking & \err & $\alpha$ & \wrg\\
      \midrule
       A: Exact (arbitrary precision)  & 0\% & -- & 0 \\
       B: Exact (small precision loss)  & 0.1\% & -- & $\leq 20$ \\
       C: Approximate (small precision loss)  & -- & 0.8 & $\leq 20$\\
      \bottomrule
    \end{tabular}
    }%
    \caption{2022/2023 Competition requirements enforced by ranking
      according to the type and declared precision in the output by
      the solver.}\label{tab:accuracy:2022ff}
  \end{subtable}
  \caption{Expected accuracy of participating solvers during the 2021--2023 Model Counting Competitions.\\ %
    $\err$~refers to the upper bound on the error, %
    $\alpha$~refers to the expected approximation factor, and %
    $\wrg$ refers to the maximum number of not accepted, i.e., wrong solutions.
  }%
  \label{tab:accuracy}
\end{table}

\newcommand{\system}[1]{\textit{#1}}

\subsection{Computing Infrastructure}\label{sec:prelims:requirements}
We follow standard guidelines for empirical
evaluations~\cite{KouweAndriesseBos18a} and measure runtime
using~\texttt{perf} and restrict runtime and memory using
\texttt{runsolver}~\cite{Roussel11}.
We execute experiments on three high performance computing (HPC)
systems consisting of the \system{StarExec}, \system{Cobra},
\system{Taurus}, and \system{Tetralith} cluster system as well as one
\system{desktop} system for which we provide additional details in
\ref{sec:appendix:platform}.
We used (a) the clusters \system{Cobra}, \system{Taurus}, and
\system{Tetralith} during the preparation phase, (b) the
\system{desktop} system for generating weighted model counting
instances, and (c) the \system{StarExec} system for the systematic
evaluation of the instances during the competition phase.
We forced the performance governors to the
base-frequency~\cite{HackenbergEtAl19a}, disabled hyper-threading, and
set transparent huge pages~\cite{kernel_thp} to always on
\system{StarExec} and kept it on the default value madvise on all
other systems.

\subsection{Submission Requirements}
We impose very weak conditions on the submissions to ease
participation.
We ask developers to submit programs that run on the Linux-based
evaluation system (\system{StarExec}) and recommend that the solvers
are provided as statically linked binary (ELF64bit) without dynamic
dependencies, including libc.
All necessary submission scripts have to be included. If the
submission uses an external third party library, which is not under an
open source license, we expect that a script for downloading the
library is included. Libraries that are not free for academic use are
not allowed.
The authors need to prepare the submission themselves by using the
StarExec virtual machine
\href{https://www.starexec.org/starexec/public/about.jsp}{\nolinkurl{image}}.
Authors may submit portfolio solvers and employ separate
pre-processors, but the program needs to announce which tool is
running in each step.
We limit the number of submissions to two per team and, in 2021 and
2022, we required that authors are part of only exactly one team.
We highly encourage solver developers to publish the source code in
the data repository and distribute with a standard open source
license~\cite{OSI_licenses}.
For archival and heritage purposes, authors need to give permission
that their solvers are uploaded to a public data repository after the
competition.

\newcommand{\bench}[1]{\texttt{#1}}
\section{Instances and Selection}\label{sec:instances}
In this section, we discuss how we collect and disseminate
instances. We provide an overview on the origin of existing instances
and elaborate characteristics of the available instances. Finally, we
give insights into the selection process of instances for the
competition.

\subsection{Instances}\label{sec:collection}
Challenging and representative benchmarks are essential to perform a
scientifically meaningful comparisons of solvers. We aim for various
large, diverse, and understandable sets of instances.
Therefore, for each iteration, we invite members of the community by
an open call for benchmarks to submit real world instances and
instance generators. Furthermore, we collecte instances from
the literature and other competitions.
For real world instances, we welcome instances that encode an
application and instances that have been obtained via a translation
from another formalism.
We decline submissions of instances that have been collected in
previous years or permuted instances thereof.
We encourage contributors to provide the instances as a dataset on the
public data repository Zenodo (\url{https://zenodo.org/}).
Since we publish all instances after the event, we expect that the
copyright of the dataset allows for publication under a CC-BY
license. If instances are sensitive (e.g.,
infrastructure/health-care), sufficient anonymization has to be
applied prior to submission.

All instances, which we collected in the process of the competition,
and instances that have been selected into each year's competition set
are publicly available in a Zenodo community, therefore follow the
link
\href{https://zenodo.org/communities/modelcounting/}{\nolinkurl{zenodo.org/communities/modelcounting}}.
We invite researchers to contribute additional instances to this
community.
For each instance set, we provide brief information about the authors,
a description of it origin, a file list including an sha256
checksum~\cite{PenardWerkhoven08a}. We compress all instances using
XZ~\cite{CollinTanothers22a} and converted instances from the 2020
competition into the recent format.
While, in 2020, instances were preprocessed, the following iterations
took instances as is without additional preprocessing. The full
instance set of the Model Counting Competition 2020 has been updated
to dedicate unpreprocessed (plain) and preprocessed (bpe/pmc)
instances.
Note that we validate the data format for each instance using a simple
program, which is publicly available~\cite{FichtePriesner23a}.
If the instance header contradicts with the data, for example, number
of variables or clauses, we reconstruct a new header from the
instance.
We did not check for duplicates among the collected instance sets.

\paragraph{Collected Instances and Competition Instances (2020)}
The 2020 instance collection consists, omitting preprocessing
instances, of in total 2,759 instances for model counting, 2,162
instances for weighted counting, and 1,106 instances for projected
model counting, which
are available on
\href{https://doi.org/10.5281/zenodo.4292167}{\nolinkurl{Zenodo:4292167}}. %
The selected competition instances can as well available on
\href{https://zenodo.org/record/3934426}{\nolinkurl{Zenodo:3934426}}. %
Details on the instances are available in the previous
report~\cite{FichteHecher20a}.
We updated all instances to the 2021 data format, corrected headers
and incomplete lines, and cleaned up the dataset. Note that this
slightly effects the total numbers of instances in each category.  The
updated dataset is available on Zenodo as most recent version.

\paragraph{Collected Instances and Competition Instances (2021)}
We additionally collected 51,333 instances, for the 2021 competition
edition. We received one set of new instances.
Instances of the full collection are available on
\href{https://doi.org/10.5281/zenodo.10006441}{\nolinkurl{Zenodo:10006441}}
and the instances that we selected for the competition are on
\href{https://doi.org/10.5281/zenodo.10012857}{\nolinkurl{Zenodo:10012857}}.
We corrected faulty headers and incomplete lines.
The contributors and origins of the instances are as follows.

\medskip
\noindent\textit{Model Counting (\MC):}\\[-1.5em]
\begin{enumerate}
\item 480 instances from the \bench{RandCluster} instance set, which
  contains challenging random instances or random instances with
  controlled variable occurence (flat cluster), contributed by
  Guillaume Escamocher and Barry O'Sullivan. A description is provided
  in the dataset.
\item 228 instances from the \bench{DIMACS-2} instance set, which
  consist of instances that have been collected during the 2nd DIMACS
  Implementation Challenge by Trick et al.~\cite{TrickChvatalCook93a}.
\item 50,294 instances from the \bench{Satlib} instance set, which we
  took from the online resource collection for research on SAT by
  Holger Hoos and Thomas Stützle~\cite{IsogaiHashimotoSakai21a};
\item 200 instances from \bench{BalancedSATComp} Marius Lindauer et
  al.~\cite{HoosKaufmannSchaub13a};
\item 131 instances from the \bench{IntervallOrderings} instance set,
  which we take from work by Sigve Hortemo Sæther, Jan Arne Telle, and
  Martin Vatshelle~\cite{SaetherTelleVatshelle15a}.
\item 11,380 instances collected by Markus Zisser from various sources
  including instances used for evaluations of \bench{ApproxMC},
  \bench{C2D}, \bench{Cachet}, \bench{DynASP} and instances collected
  by Daniel Fremont and of the \bench{Satlib} set, and instances
  collected earlier that were used in the evaluation of
  gpusat~\cite{FichteEtAl18c,FichteHecherZisser19a}.
  Note that this set might introduce duplicates.
\end{enumerate}

\medskip
\noindent\textit{Weighted Model Counting (\WMC).}
We received no new weighted instances. To still increase our instance
stock, we introduced two instance generators, which we briefly
describe in %
\ref{sec:appendix:generator}. The generators take existing model
counting instances and generate weights. The first generator randomly
sets weights in the given input instance. The second program takes the
given CNF instance and constructs a Decision-DNNF (decision
decomposable negation normal form) using a knowledge compiler, then
generates the corresponding counting
graph~\cite{Darwiche01a,Darwiche04a}, and finally guesses normalized
weights along the counting graph.

\medskip
\noindent\textit{Projected Model Counting (\PMC).}
We were not presented with submissions of instances from new domains.
Nonetheless, to obtain challenging new instances for the next
iteration, we again take previously collected instances and randomly
guess projection variables.
Brief notes on the implementation can be found in
\ref{sec:appendix:generator}.

\paragraph{Collected Instances and Competition Instances (2022)}
We received in total 7,732 instances that originate from 5 domains.
All collected instances are publicly available on
\href{https://doi.org/10.5281/zenodo.10014715}{\nolinkurl{Zenodo:10014715}}.
The instances that we selected for the competition are available on
\href{https://doi.org/10.5281/zenodo.10012860}{\nolinkurl{Zenodo:10012860}}.
We describe the selection process in more details in
Section~\ref{sec:instances:selection}.
The contributors and origins of the instances are as follows.

\medskip\noindent\textit{Model Counting (\MC):}
\begin{enumerate}
\item 1,139 instances from the set called \bench{Argumentation}, which
  encode finding the count to abstract argumentation frameworks for
  various
  semantics~\cite{Dung95a,LagniezLoncaMailly20a,NiskanenJarvisalo20a}. The
  instances were contributed by Piotr Jerzy Gorczyca.
\item 41 instances from the \bench{Inductive Inference} instance set,
  where Boolean function synthesis problem are encoded into
  propositional
  satisfiability~\cite{KamathKarmarkarRamakrishnan92a}. Yong Lai
  submitted the instance set.
\item 27 instances from a set called \bench{coloring} contributed by
  Daniel Pehoushek that encode counting graph colorings into a model
  counting task.
\item 6,253 instances from a set called \bench{AspNavigation}
  contributed by Dominik Rusovac~\cite{FichteGagglRusovac22}, which
  asks for counting the number of answers sets under assumptions
  encoded into propositional
  formulas~\cite{FichteGagglHecher22a,Janhunen04a,Bomanson17a}.
\item 22 instances from a \bench{Sudoku} benchmark set, which encodes
  the well-known Sudoku puzzle into CNF
  instances~\cite{LynceOuaknine06a,PfeifferKarnagelScheffler13a}.
  Since a Sudoku puzzle technically only contains exactly one
  solution, some initial values are removed to increase the number of
  satisfying assignments.
  The instances were contributed by Ivor Spence~\cite{Spence22a};
\item 127 instances from \bench{Industrial Feature Models}, which
  consider families of products in industry that share multiple
  configuration options resulting in large configuration spaces. The
  instances encode computing the number of valid configurations for
  tasks such as estimating the effort of an update or effectively
  reducing the configuration
  space~\cite{SundermannThumSchaefer20}. The instances were provided
  by Chico Sundermann, Thomas Thüm, and Ina Schaefer.
\end{enumerate}

\medskip
\noindent\textit{Weighted Model Counting (\WMC):}
As in 2021, we received no new weighted instances. Again, we secured
additional instances by employing our instance generator. We refer to
\ref{sec:appendix:generator}.

\medskip
\noindent\textit{Projected Model Counting (\PMC):}
\begin{enumerate}
\item 123 instances from \bench{Software Reliability Quantification},
  which encode reliability questions, meaning, comparing C programs
  against a given functional specification, and computing the
  conditional probability of violating an assertion, i.e., count under
  the violating assertion related to all terminating
  runs~\cite{TeuberWeigl21}.
  The instances were contributed by Samuel Teuber and Alexander
  Weigl~\cite{TeuberWeigl22a}.
\end{enumerate}
To obtain additional instances, we again take existing instances and
randomly guess projection variables.

\paragraph{Collected Instances and Competition Instances (2023)}
We received three new sets of model counting instances.
Overall, we collected 840 instances for the 2023 competition edition,
which are publicly available on
\href{https://doi.org/10.5281/zenodo.10012822}{\nolinkurl{Zenodo:10012822}}.
The instances that we selected for the competition are on
\href{https://doi.org/10.5281/zenodo.10012864}{\nolinkurl{Zenodo:10012864}}.
We provide details for the instance selection process in
Section~\ref{sec:instances:selection}.
The contributors and origins of the instances are as follows.

\medskip
\noindent\textit{Model Counting (\MC):}
\begin{enumerate}
\item 191 instances from a \bench{Network Reliability} instance set,
  which contains encoded network reliability estimation on real world
  power grids~\cite{KabirMeel23a}. Mohimenul Kebir and Kuldeep Meel
  submitted these instances.
\item 643 instances from the \bench{BitVectorCMT} instance set, which
  consists of CNFs that encode problems for bitvector counters. The
  instances were generated using the ABC
  system~\cite{FanWu23a,KimMcCaman20a,FanWu23b} and provided by Arijit
  Shaw and Kuldeep Meel.
\item 6 instances from \bench{Path Counting}, which encodes to count
  the number of paths through a rectangular grid of cells with
  additional constraints. The description of the instance set, which
  is provided along the instance set, contains additional details. The
  instances together with a generator were provided by Ivor Spence.
\item 56 instances from \bench{Robotic Planning}, which encodes a
  robotic planning problem that was initially encoded using Z3's
  theory of bitvectors and then bitblasted down to SAT. The instances
  were provided by Eric Vin~\cite{GittisVinFremont22}.
\end{enumerate}

\medskip
\noindent\textit{Weighted Model Counting (\WMC):}
As in the 2020 and 2021 editions, we received no new weighted
instances. Again, we added instances by employing our instance
generator, see~\ref{sec:appendix:generator}.

\medskip
\noindent\textit{Projected Model Counting (\PMC):}
We did not receive new instances for projected model counting.  Again,
we take previously collected instances from all categories and
randomly guess projection variables. We refer to
\ref{sec:appendix:generator}.

\subsection{Selection of Instances}\label{sec:instances:selection}
From the collected instances as stated in
Section~\ref{sec:collection}, we selected 200 instances for each
track.
We numbered the instances from 1 to 200 and selected the odd numbered
instances as private and even numbered instances as public
instances. We disclosed the 100 public instances publicly at
\href{https://mccompetition.org/}{\nolinkurl{mccompetition.org}}
during the submission and testing phase of the competition in May each
year.
Shortly before the presentation of the results at the SAT conference,
we disclosed the 100 private instances to the participants.
Then, in early fall, we published the set of all instances publicly for
download at the model counting website.
Finally, we compiled Zenodo repositories when preparing the report.
During the 2021--2023 editions of the competitions, we followed two
different approaches as outlined below.

\paragraph{2021 Competition (Balanced)}
For the 2021 Competition, we decided to pre-select instances according
to expected practical hardness and similar to the process in 2020. Our
underlying idea was to pick instances that are challenging, vary in
size, are up to a certain extend still within reach for the
participants and challenge them.
We aim for a stable benchmark set where we can observe useful
differences between solvers and that might later also provide
interesting insights on the progress of model counting.
However, we would also like to mention that this selection process may
cause problems such as an incalculable advantage for existing
techniques, massive precomputation time on various solvers, and hence
an unnecessary amount of work in the preparation phase of the
competition.
In order to classify the instances, we ran existing exact solvers from
the previous iteration on all instances with a timeout of 2 hours.
The solvers included: \solver{addmc}, \solver{c2d}, \solver{D4},
\solver{GANAK}, and \solver{sharpSAT}.
We ensured that if an instance could be solved only by one solver that
we did not pick more than 10 such instances.
We selected 40 instances that could not be solved within 3600s and
picked the remaining ones randomly from 20 instances that could be
solved within 60s, 20 within the interval (60s,250s], 20 within the
interval (250s,500s], 50 within the interval (500s,900s], 30 within
the interval (900s,1800s], and 20 within the interval (1800s,3600s] if
an interval could not be filled we picked the remaining ones randomly.

\paragraph{2022--23 (Random by Benchmark Set)}
Following suggestions from other competitions, developers, and the
technical advisor, we changed the process to select a certain number
of instances randomly by benchmark set in 2022--2023.
First, we remove instances from consideration that can be solved by
the original version of the \solver{SharpSAT} solver in 60s and
discard unsatisfiable instances due to limited interest for the actual
counting task.
Then, we randomly sample 3 unsolved and 12 solved instances per
benchmark set. If there are fewer instances, we take all instances.
We randomly sample 160 solved and 40 unsolved instances from this
selection.
Our primary idea is to avoid a strong bias to large or small benchmark
sets and incorporate instances that could have previously not been
solved.

\paragraph{Preprocessing}%
While there are preprocessing techniques available for various
versions of model counting, we did not apply any preprocessing since
2021 to the instances and left this entirely to the developers of the
solvers.
Recent preprocessors include \solver{pmc}~\cite{LagniezMarquis14a},
\solver{B+E}~\cite{LagniezLoncaMarquis20}, and
\solver{Arjun}~\cite{SoosMeel22}.
For model counting, documented options for the \solver{pmc}
preprocessor are \ops{-vivification} \ops{-eliminateLit}
\ops{-litImplied} \ops{-iterate=10} \ops{-equiv} \ops{-orGate}
\ops{-affine}.
Documented options for \solver{B+E} are~\ops{-limSolver=1000}.
For projected model counting, \solver{pmc} can be used as preprocessor
using the options \ops{-vivification} \ops{-eliminateLit}
\ops{-litImplied} \ops{-iterate=10}

\paragraph{Precomputation of Model Counts}
To precompute the model counts, we ran the three best counters from
the previous iteration on the selected instances.  The solvers
included \solver{addmc}, \solver{ApproxMC}, \solver{c2d}, \solver{D4},
\solver{DPMC}, \solver{GANAK}, and \solver{GPMC}. Note that we
consider non-exact solvers only to confirm the order of previously
computed model counts.

\begin{table}[tb]
  \centering
  \begin{tabular}{HlHHHrrHrrrrr}
    \toprule
       & set    & public/private & solved & twub & \#  & twub[Mdn] & twub[max]  & n[Mdn]  & n[max]  & m[Mdn] & c[Mdn]  & log10[Mdn] \\
    \midrule
8 & 2021 & private & -- & + & 10 & 435 & 507,949 & 3,231 & 101,451 & 76,694 & 18 & -- \\
    28 & 2022   & private        & --     & +    & 17  & 195       & 507,949.00 & 729     & 50,461  & 10,543 & 18      & --       \\
    48 & 2023   & private        & --     & +    & 10  & 172       & 77,348.00  & 3,834   & 130,982 & 26,873 & 40      & --       \\
    \midrule
9 & 2021 & private & + & + & 90 & 56 & 737 & 1,494 & 119,057 & 5,396 & 36 & 21 \\
    29 & 2022   & private        & +      & +    & 83  & 76        & 964.00     & 759     & 349,990 & 3,174  & 20      & 14       \\
    49 & 2023   & private        & +      & +    & 85  & 67        & 631,270.00 & 3,754   & 158,705 & 18,764 & 66      & 16       \\
    50 & 2023   & private        & +      & --   & 5   & --        & --         & 137,358 & 217,363 & 82,069 & 129,614 & 39,088   \\
    \bottomrule
 \end{tabular}
 \caption{Characteristics of instances from Track 1 (MC) grouped by unsolved and solved. %
   ``\#'' refers to number of instances,
   ``twub'' to an upper bound of the treewidth of the primal
   graph,
   ``[Mdn]''  to the median, ``n'' to the number of variables,
   ``m'' to the number of clauses, ``log10'' to the model count in
   log10-notion, and ``--'' means that the instances are unsolved or
   we could not compute an upper bound on the treewidth.
 }
 \label{tab:track1:character}
\end{table}

\paragraph{Instance Characteristics}
Table~\ref{tab:track1:character} briefly summarizes instance
characteristics of the competition instances from Track~1 (MC).
We omit statistics for the weighted, projected, and weighted projected
tracks to avoid false insights. Recall that we obtained numerous
instances by taking model counting instances and guessing weights or
projection variables.
We consider the number of solved instances of the virtual best solver,
an upper bound on the treewidth (median), number of variables
(median/max), number of clauses (median), number of communities, and
model count (median).
We computed the upper bound on the treewidth using
flowcutter~\cite{HamannStrasser18} and htd~\cite{AbseherMusliuWoltran17a}.
The number of communities using tools provided by Ansótegui,
Giráldez-Cru, and Levy~\cite{AnsoteguiGiraldez-CruLevy12}.
We observe that the overall number of unsolved instances declines from
2022 to 2023. For unsolved instances, the median on a treewidth upper
bound appears to be quite high.
The median over the number of variables and number of clauses
increases (759, 3,754). Surprisingly, the maximum on the number of
variables (349,990) and number of clauses (82,069) can be much higher
than one would expect for counting from an uneducated theoretical
perspective.
Unsurprisingly, the median over the log10 of model counts is much
higher than one could possibly carry out by enumerating solutions.
To investigate possible correlations with solving, we investigated
community structures, modularity~\cite{AnsoteguiGiraldez-CruLevy12},
and treewidth.
So far, we cannot identify a correlation for community structures and
modularity. For upper bounds on the treewidth, it appears that we can
find only higher upper bounds on the width, which however does not
give a correlation as we heuristically compute the width.
Since we selected instances in 2022 and 2023 randomly, we cannot draw
conclusions from that fact that the solvers manage to solve more
instances.
Finally, this leaves us with a research question: what are good
measures to characterize model counters -- preferably models that also
incorporate the underlying methods and benchmark sets.

\begin{figure}
  \centering
  \includegraphics{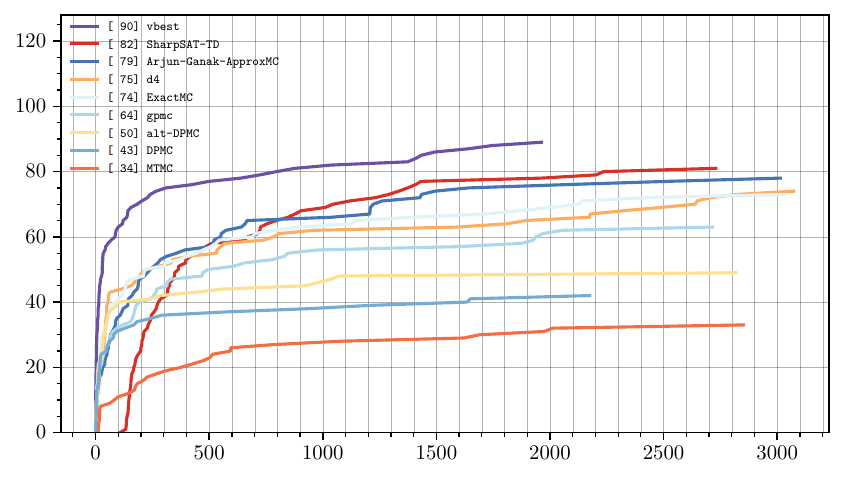}
  \caption{2023 Iteration Track 1 (MC). Runtime results illustrated as
    cumulated solved instances. The y-axis labels consecutive integers
    that identify instances. The x-axis depicts the runtime. The
    instances are ordered by running time, individually for each
    solver.}
  \label{fig:mc2023:track1:cdf}
\end{figure}

\section{Participants and Results}\label{sec:results}
Throughout the three years of the competition we had a quite stable
number of participants.
For \mbox{Track 1~(MC)}, we received 10 submissions in 2021, 11 submissions
in 2022, and 9 submissions in 2023 showing a small decline in
submissions.
For Track 2~(WMC), we received 6 submissions in 2021, 5 submissions in
2022, and 7 submissions in 2023. Clearly, developers of the leading
solvers took time to enhance the solver by weighted counting
capabilities.
For Track 3~(PMC), we received 5 submissions in 2021, 4 submissions in
2022, and 7 submissions in 2023 showing that we now also have a good
variety of solvers that support projected counting.
For Track 4~(PWMC), we received 2 submissions in 2022 and 4
submissions in 2023. Thus, we can see a small increase in solvers that
support now projection in connection with weights.
We received no submission in the heuristic track.
Our developers came from China, Finland, France, Germany, India,
Japan, Poland, Norway, Singapore, UK, and USA.
In the following, we only state the names of the authors for the
leading submissions. For detailed information on the team members of
the submissions, we refer to \ref{appendix:submissions}.
All submissions are publicly available on %
  \href{https://zenodo.org/records/10012811}{Zenodo:10012811 (2023)},
  \href{https://zenodo.org/records/10012803}{Zenodo:10012803 (2022)},
  \href{https://zenodo.org/records/10006718}{Zenodo:10006718 (2021)}, and 
  \href{https://zenodo.org/records/10029900}{Zenodo:10029900 (2020)}.
Recall that Table~\ref{tab:mc-solvers} also surveys the original
developers, sources, and academic reference of participating solvers.
We would like to point out that we evaluate different rankings on the
same instances to avoid additional and precious computation time.
Due to the amount of resources, tracks, and iterations, we focus on
the number of solved instances and how scores developed over the
years.
However, we hope that some readers also appreciate the additional
analysis, which we provide publicly on Zenodo at
\href{https://zenodo.org/records/10671987}{Zenodo:10671987 (StarExec Data \& Evaluations)}.
There we report on all executed solver configurations including
runtime, model count (log10-notation and exact counts), comparison of
computed counts by solvers and instances, precision, accuracy, error,
expected counts, or wrong results.
These details have been shared with the developers prior to the
presentations of results at the SAT conferences.

\begin{table}[t!]
  \centering
  \begin{subtable}[t]{0.48\textwidth}
    \resizebox{.95\textwidth}{!}{%
  \begin{tabular}{c@{\hspace{0.5ex}}c@{\hspace{0.5ex}}clHcc}
    \toprule
    0\% & 0.1\% & 0.8 & Submission           & Authors                                                                 & solved \\
    \midrule
    1   & 1     & 1   & \solver{SharpSAT-TD}          & Tuukka Korhonen, Matti Järvisalo                                        & 82     \\
        &       & 2   & \solver{Arjun-GANAK-ApproxMC} & Mate Soos, Kuldeep S. Meel                                              & 79     \\
    2   & 2     & 3   & \solver{D4}                   & Pierre Marquis, Jean-Marie Lagniez                                      & 75     \\
        & 3     & 4   & \solver{ExactMC-Arjun}        & Yong Lai, Zhenghang Xu, Minghao Yin,  Kuldeep S. Meel,  Roland H.C. Yap & 74     \\
        & 4     & 5   & \solver{Arjun-GANAK}          & Mate  Soo,  Kuldeep S. Meel                                             & 71     \\
    3   & 5     & 6   & \solver{GPMC}                 & Kenji Hashimoto,  Shota Yap                                             & 64     \\
        & 6     & 7   & \solver{Alt-DPMC}             & Aditya Shrotr,  Moshe Vardi                                             & 50     \\
    4   & 7     & 8   & \solver{DPMC}                 & Vu Phan, Jeffrey Dudek, Moshe Vardi                                     & 43     \\
        & 8     & 9   & \solver{mtmc}                 & Ivor Spence                                                             & 34     \\
    \bottomrule\\
  \end{tabular}
  }%
  \caption{Track 1 (MC): 2023 Iteration}
  \end{subtable}
  \begin{subtable}[t]{0.48\textwidth}
    \resizebox{.95\textwidth}{!}{%
  \begin{tabular}{c@{\hspace{0.5ex}}c@{\hspace{0.5ex}}cp{10em}Hc}
    \toprule
    0\% & 0.1\% & 0.8 & Submission                 & Authors                                                  & solved \\
    \midrule
    1   & 1     &     & \solver{SharpSAT-TD-Arjun}          & Mate Soos Kuldeep S. Meel                                & 79     \\
    2   & 2     &     & \solver{ExactMC}                    & Yong Lai, Kuldeep S. Meel. Roland H.C. Yap, Zhenghang Xu & 77     \\
    3    &  3     &     & \solver{SharpSAT-TD}                & Tuukka Korhonen, Matti Järvisalo                         & 77     \\
    4   & 4     &     & \solver{D4}                         & Pierre Marquis, Jean-Marie Lagniez                       & 76     \\
        &       & 1   & \solver{SharpSAT-TD-Arjun-Appr} & Mate Soos, Kuldeep S. Meel                               & 74     \\
    5   & 5     &     & \solver{GPMC}                       & Kenji Hashimoto, Shota Yap                               & 69     \\
    6   & 6     &     & \solver{mtmc}                       & Ivor Spence                                              & 66     \\
    7   & 7     &     & \solver{DPMC}                       & Vu Phan, Jeffrey Dudek, Moshe Vardi                      & 61     \\
        & 8     &     & \solver{TwG}                        & Sylvester Swats                                          & 53     \\
    8   & 9     &     & \solver{c2d}                        & Adnan Darwiche                                           & 50     \\
    \bottomrule
  \end{tabular}
  }%
  \caption{Track 1 (MC): 2022 Iteration}
  \end{subtable}
  \\[1em]%
  \begin{subtable}[t]{0.48\textwidth}
    \centering
  \begin{tabular}{llHlHH}
    \toprule
    \# & Submission    & Authors                                      & solved & excl \\
    \midrule
    1  & \solver{SharpSAT-TD}   & Tuukka Korhonen,  Matti Järvisalo            & 78     & 0    \\
    2  & \solver{narasimha} & Sharma,  Lai, Xu,  Roy, Yap, Mate Soos, Meel & 61     & 1    \\ %
    3  & \solver{D4}            & Jean-Marie Lagniez,  Pierre Marquis          & 51     & 0    \\
    4  & \solver{GPMC}          & Kenji Hashimoto, Takaaki Isogai              & 38     & 0    \\
    5  & \solver{MC2021\_swats} & Sylwester Swat                               & 34     & 0    \\
       & \solver{DPMC}          & Vu Phan, Jeffrey Dudek, Moshe Vardi          & 34     & 0    \\
    7  & \solver{c2d}           & Adnan Darwiche                               & 29     & 0    \\
    8  & \solver{bob}           & Daniel Pehoushek                             & 11     & 0    \\
    9  & \solver{SUMC2}         & Ivor Spence                                  & 7      & 0    \\
    \bottomrule
  \end{tabular}
  \caption{Track 1 (MC): 2021 Iteration}
  \end{subtable}
  \begin{subtable}[t]{0.48\textwidth}
    \centering
    \begin{tabular}{llHlH}
      \toprule
      \# & Submission           & From                         & solved & excl \\
      \midrule
      1 & \solver{SharpSAT-TD}          & Helsinki                     & 68     & 0    \\
      2 & \solver{narasimha}         & Singapore, Kanpur, Changchun & 65     & 0    \\
      3 & \solver{D4}             & Lens                         & 53     & 0    \\
      4 & \solver{c2d}         & LA                           & 50     & 0    \\
      5 & \solver{DPMC}          & Huston                       & 48     & 0    \\
      6 & \solver{MC2021\_swats}         & Poznan                       & 40     & 0    \\
      7 & \solver{SUMC2}                & Belfast                      & 26     & 0   \\
      \bottomrule\\\\
    \end{tabular}
    \caption{Track 4 (MC): 2021 Iteration}
  \end{subtable}
  \caption{Results Track 1 (MC). Number of solved private instances out of 100. Table~(d) contains an additional set 
    of model counting instances (2021 Track~4), which was evaluated on less restrictive requirements for precision.
  } %
  \label{tab:mc}
\end{table}

\subsection{Solvers}

Before we discuss the results, we provide brief descriptions of
prominent solvers
and involved tools as directly provided by the developers or
summarized by us. Descriptions directly given by the
authors %
are given in quotation marks (``[...]'') and typesetted in italics.

\paragraph{\solver{Arjun-GANAK-ApproxMC}}
\noindent Mate Soos describes the solvers by him, Kuldeep S. Meel, his
group, and collaborators in more extend online~\cite{Soos24}.
Their submissions to the competition consisted primarily of manually
time-based portfolios where one solver or preprocessor runs for a
particular time, if no solution was found another solver continues
\solver{ApproxMC}~\cite{SoosMeel19} is a probabilistically approximate
model counter, which is based on partitioning the search space using
XOR constraints and decision-based approximation. The decisions are
done by CryptoMiniSat~\cite{SoosNohlCastelluccia09}, which is a SAT
solver by Mate Soos that supports Gaussian Elimination on XOR
constraints.
\solver{Arjun}~\cite{SoosMeel22} is preprocessor that computes a
subset of projection variables such that these variables are
independent support, i.e., whenever models restricted to independent
support variables are identical, they are identical identical on all
projection variables, which enables fast preprocessing.
\solver{GANAK}~\cite{SharmaEtAl19a} is a probabilistic exact model
counter based on \solver{SharpSAT}, \solver{SharpSAT-td}, and
\solver{GMPC}. Its core is a component-caching-based solving approach
with probabilistic component caching, enhanced branching heuristics,
and phase selection heuristics.

\paragraph{\solver{c2d}} The \solver{c2d} knowledge compiler is one of
the early tools for model counting developed and submitted by Adnan
Darwiche~\cite{Darwiche04a}. It compiles a formula in CNF a into
d-DNNF (Deterministic Decomposable Negation Normal Forma). A d-DNNF
permits model counting in polynomial time. The implementation employs
techniques from SAT and OBDD solving.

\paragraph{\solver{D4}} The \solver{D4} knowledge compiler employs
top-down search taking advantage of dynamic decompositions based on
hypergraph partitioning~\cite{LagniezMarquis17a}. In the recent
versions,  preprocessing techniques have been extended.

\paragraph{\solver{DPMC}}
The \solver{DPMC} solver is based on dynamic-programming employing
tree decompositions (project-join orders) and efficient data
structures,~e.g., algebraic decision diagrams with
tensors~\cite{DudekPhanVardi20b}.

\paragraph{\solver{GPMC}}

\noindent Ryosuke Suzuki, Kenji Hashimoto, and Masahiko
Sakai~\cite{SuzukiHashimotoSakai17a} write about their \solver{GPMC}
solver: %
{\itshape%
  ``We improve our existing projected model-counting solver with
  component decomposition and caching. We use a CDCL algorithm,
  including backjumping and restarting, to decide whether components
  without projection variables are satisfiable or not. Furthermore, we
  try a limited backjumping technique according to the condition of
  processed components. We reimplement, on glucose, a projected model
  counter with component decomposition and caching. Then we add the
  above functions and evaluate by experiments them comparing with the
  existing projected model counters.''
}

\paragraph{\solver{ExactMC}}
The \solver{ExactMC} solver~\cite{LaiMeelYap21a} implements model
counting by knowledge compilation based on a generalization of
Decision-DNNF, called CCDD, which enable literal equivalences.
Versions of the solver also allow for anytime counting without
guarantees,~i.e., only providing bounds if the solver is interrupted.

\paragraph{\solver{mtmc}} %
\noindent Ivor Spence provided us with the following short
description:
{\itshape%
``The Merge Tree Model Counter (mtmc) solves the model counting
(including projected and weighted versions) using a static variable
ordering. An explicit binary tree (actually a directed acyclic graph)
is constructed to represent and then count the satisfying models. The
idea is most effective when a good variable ordering can be found to
reduce the range of variables within clauses, either by simulated
annealing or by a "force" algorithm. The order in which clauses are
added as the tree grows to represent the entire problem is also
important. The solver typically uses more memory than others.''
}

\paragraph{\solver{ProCount}}
The \solver{ProCount} solver~\cite{DudekPhanVardi21a} employs
dynamic-programming on tree decompositions (project-join trees)
Central concept is to define gradedness on project-join trees, which
require irrelevant variables to be eliminated before relevant
variables.

\paragraph{\solver{narasimha}}
Is a portfolio consisting of \solver{ApproxMC}, \solver{B+E}
preprocessing, \solver{ExactMC}, and \solver{Ganak} first
preprocessing runs, then \solver{Ganak}, then \solver{ExactMC} or
\solver{ApproxMC}.

\paragraph{\solver{SUMC2}}
Ivor Spence~\cite{Spence22} describes the \solver{SUMC2} solver as
follows: {\itshape%
  \solver{SUMC2} is new version of the \solver{SUMC1} solver written
  specifically to test an idea about how to count the number of models
  which satisfy a propositional expression. \solver{SUMC1} was written
  to enter into the first model counting
  competition~\cite{FichteHecherHamiti21a}. The solver is based around
  counting how many models are eliminated by each clause because they
  fail to satisfy it, and using an extended mathematical result giving
  the cardinality of a union of sets to determine how many models are
  eliminated overall. The GNU Multiple Precision Arithmetic Library
  (GMP~\cite{GranlundEtAl16a}) is used to manipulate the large integer
  values which arise because the possible number of models is so
  large.  The solver calculates exactly how many satisfying models
  there are but overall its performance is not expected to be
  competitive with the best other solvers when considered across the
  full range of benchmarks, in particular in an environment in which
  memory is constrained. It is however possible that it will perform
  well on small benchmarks which other solvers may find difficult.
  The source code of the solver is available at
  \url{https://www.github.com/ivor-spence/sumc}.
}%

\paragraph{\solver{SharpSAT-TD} (Tuukka Korhonen, Matti J\"arvisalo)}
\noindent Tuukka Korhonen and Matti
J\"arvisalo~\cite{KorhonenJarvisalo23a} write in their solver
description for the competition version of the solver:
{\itshape%
  ``SharpSAT-TD is based on the exact model counter
  SharpSAT~\cite{Thurley06a}, from which it inherits the basic
  structure of a search-based model counter with clause learning,
  component analysis, and component caching. The main new feature in
  SharpSAT-TD is that we compute a tree decomposition of the input
  formula with the FlowCutter
  algorithm~\cite{HamannStrasser18,Strasser17}, and integrate the tree
  decomposition to the variable selection heuristic of the counter by
  a method introduced by the authors
  in~\cite{KorhonenJarvisalo21a}. Another significant new feature is a
  new preprocessor. Further, SharpSAT-TD extends SharpSAT by directly
  supporting weighted model counting.
  [...]
  We note that the current version of SharpSAT-TD differs
  significantly from the version evaluated
  in~\cite{KorhonenJarvisalo21a}, as the version evaluated
  in~\cite{KorhonenJarvisalo21a} differed from SharpSAT only in the
  variable selection heuristic, while the current SharpSAT-TD has also
  other new features. The current SharpSAT-TD has stayed almost
  unchanged since the Model Counting Competition 2021: After the
  competition we fixed some bugs and added proper support for weighted
  model counting, but after that no further updates have been made.''
}

\subsection{Track 1 (MC)}\label{sec:results:mc}
Table~\ref{tab:mc} gives a detailed overview on the standings and
solvers in the years 2021, 2022, and 2023.
Figure~\ref{fig:mc2023:track1:cdf} illustrates runtime results in form
of a CDF for the 2023 iteration. Note that we technically only borrow
the visual representation and underlying idea of CDFs (cumulative
distribution functions), which describe the distribution of random
variables by summing up the probability of an event smaller than a
specific value -- in our case the runtime.
We employ the ranking as given in Section~\ref{sec:ranking}.
Overall, we have three extremely strong teams and techniques, namely
knowledge compilation, component-based caching, and preprocessing.

In 2021, a new entrant entered the field in model counting. Tuukka
Korhonen and Matti Järvisalo developed a solver named
\solver{SharpSAT-TD}~\cite{KorhonenJarvisalo23a,KorhonenJarvisalo21a},
which builds on a well-established solver \solver{SharpSAT} by Marc
Thurley~\cite{Thurley06a} that uses component-caching. Tukka and Matti
modified the heuristic by enhancing it using tree decompositions and
improved preprocessing. \solver{SharpSAT-TD} solved 17 instances more
than the next best solver. This was a huge and unexpected advancement.
The solver turned out to be the new baseline in the coming years.
The runner-up was narasimha by Sharma, Lai, Xu, Roy, Yap, Soos, and
Meel, who combined approximate
(\solver{ApproxMC}~\cite{ChakrabortyMeelVardi16a}) and exact solving
(\solver{GANAK}~\cite{SharmaEtAl19a}) into a portfolio and solved 61
instances. The third place is filled by
\solver{D4}~\cite{LagniezMarquis17a} submitted by Jean-Marie Lagniez
and Pierre Marquis, who based their solver on knowledge compilation,
and solved 51 instances.
In summary, we can see a huge gap between the three best solvers.
Note that we combined approximate and exact solving techniques in the
ranking Track~1 (MC) in 2021, which results in the fact that a few
instances were outside of expected margin of error. To compensate, we
added another track on model counting (Track~4) where we allowed a
higher margin of error. On Track~4 (MC), we observe that
\solver{SharpSAT-TD} solves in total 68 instances, closely followed by
\solver{narasimha} with 65 solved instances, and \solver{D4} with 53
solved instances. The approximate component of \solver{narasimha}
helped to solve more instances than \solver{D4}. 
In 2022, a modified version of \solver{SharpSAT-TD} together with a
fast preprocessor, called \solver{Arjun}~\cite{SoosMeel22}, manged to
solve 79 instances. It scored only two instances more than runner-ups
\solver{ExactMC}~\cite{LaiMeelYap21a} and \solver{SharpSAT-TD}, which
solved 77 instances. \solver{ExactMC} was developed by Yong Lai,
Kuldeep S. Meel.  Roland H.C. Yap, and Zhenghang Xu and is based on
knowledge compilation using CCDD (Constrained Conjunction and Decision
Diagram)~\cite{LaiMeelYap21a}.
\solver{SharpSAT-TD} was developed by Tuukka Korhonen and Matti
Järvisalo, who carried out only minor improvements over its
predecessor.
These solvers were closely followed by \solver{D4} developed by Pierre
Marquis and Jean-Marie Lagniez, who managed to solve 76 instances.
The leading submission that employed approximate solving was a
portfolio developed by Mate Soos and Kuldeep S. Meel, who managed to
solved 74 instances. Overall, there is only a small gap between the 5
best solvers or combinations of solvers.
In 2023, we again saw that \solver{SharpSAT-TD} by Tuukka Korhonen and
Matti Järvisalo was ahead of all other solvers with 82 solved
instances.
The combination of \solver{Arjun-GANAK-ApproxMC} solved 79 instances
and ranked second in the approximate ranking.
\solver{D4} again by Pierre Marquis and Jean-Marie Lagniez solved 75.
The combination of \solver{ExactMC} and Arjun by Yong Lai, Zhenghang
Xu, Minghao Yin, Kuldeep S. Meel, and Roland H.C. Yap solved 74
instances.
Closely followed, we see \solver{GANAK} together the \solver{Arjun}
preprocessor submitted by Mate Soos and Kuldeep S. Meel with 71 solved
instances.
We suspect that the preprocessors such as \solver{Arjun} played a
central role in the overall good performance of the best solvers.

\begin{table}[t]
  \centering
  \begin{subtable}[t]{0.48\textwidth}
    \centering
  \begin{tabular}{cclHcc}
    \toprule
    0\% & 0.1\% & Submission    & Authors                                                                & solved        \\
    \midrule
    1   & 1     & \solver{SharpSAT-TD}   & Tuukka Korhonen, Matti Järvisalo                                       & 75            \\
        & 2     & \solver{D4}            & Pierre Marquis, Jean-Marie Lagniez                                     & 67            \\
    2   & 3     & \solver{GPMC}          & Kenji Hashimoto, Shota Yap                                             & 62            \\
    3   & 4     & \solver{ExactMC-Arjun} & Yong Lai, Zhenghang Xu, Minghao Yin,Kuldeep S. Meel,  Roland H.C. Yap, & 59            \\
        & 5     & \solver{Alt-DPMC}      & Aditya Shrotri, Moshe Vardi                                            & 36            \\
        & 6     & \solver{DPMC}          & Vu Phan, Jeffrey Dudek, Moshe Vardi                                    & 25            \\
        & 7     & \solver{mtmc}          & Ivor Spence                                                            & 20            \\
    \bottomrule
  \end{tabular}
  \caption{Track 2 (WMC) 2023 Iteration}
  \end{subtable}
  \begin{subtable}[t]{0.48\textwidth}
    \centering
  \begin{tabular}{cclHll}
    \toprule
    0\% & 0.1\% & Submission    & Authors                                                                & solved        \\
    \midrule
    1   & 1     & \solver{SharpSAT-TD}   & Tuukka Korhonen, Matti Järvisalo                                       & 75            \\
        & 2     & \solver{GPMC}          & Kenji Hashimoto, Shota Yap                                             & 68            \\
        & 3     & \solver{D4}            & Pierre Marquis, Jean-Marie Lagniez                                     & 66            \\
    2   & 4     & \solver{c2d}           & Adnan Darwiche                                                         & 60            \\
    3   & 5     & \solver{DPMC}          & Vu Phan, Jeffrey Dudek, Moshe Vardi                                    & 43            \\
    \bottomrule\\\\
  \end{tabular}
  \caption{Track 2 (WMC) 2022 Iteration}
  \end{subtable}                                                                                                         \\[1em]
  \begin{subtable}[t]{0.48\textwidth}
    \centering
  \begin{tabular}{cHlHcc}
    \toprule
    \#  &       & Submission    & Authors                                                                & solved & excl \\
    \midrule
    1   & 1     & \solver{SharpSAT-TD}   & Tuukka Korhonen, Matti Järvisalo                                       & 90     & 1    \\
    2   &       & \solver{D4}            & Jean-Marie Lagniez, Pierre Marquis                                     & 80     & 0    \\
    3   &       & \solver{c2d}           & Adnan Darwiche                                                         & 79     & 0    \\
    4   &       & \solver{DPMC}          & Vu Phan, Jeffrey Dudek Moshe Vardi                                     & 46     & 0    \\
    5   &       & \solver{narsimha}  & Sharma,  Lai, Xu,  Roy, Yap,  Soos,  Meel                              & 25     & 37   \\
    \bottomrule
  \end{tabular}
  \caption{Track 2 (WMC) 2021 Iteration}
  \end{subtable}
  \caption{Results Track 2 (WMC)}
  \label{tab:summary:wmc}
\end{table}

\subsection{Track 2 (WMC)}\label{sec:results:wmc}
Table~\ref{tab:summary:wmc} gives a detailed overview on the
standings and solvers in the years 2021, 2022, and 2023 according to
the rankings in Section~\ref{sec:ranking}.
In 2021, Tuukka Korhonen and Matti J\"arvisalo also submitted their
\solver{SharpSAT-TD} solver to the weighted model counting track and
immediately ranked first with 90 solved instances. They outputted one
instance, which however turned out to be buggy. According to the 2021
rules, the instance was only excluded from the ranking without
disqualifying the solver. Overall \solver{SharpSAT-TD} solved 10
instances more over the second best solver.
In fact, the runner-up was \solver{D4} developed by Jean-Marie Lagniez
and Pierre Marquis who managed to solve 80 instances. Closely
followed, we have the \solver{c2d} solver with 79 solved instances
submitted and developed by Adnan Darwiche.
In 2022, the picture was almost similar, however, with a new entrant
in-between.
Since the 2022 ranking allows to distinguish between exact together
with arbitrary precision (0\%) and exact but small precision loss
(0.1\%), we see a slightly different picture.
\solver{SharpSAT-TD} by Tuukka Korhonen and Matti Järvisalo is the
clear winner regardless of using arbitrary precision with 75 solved
instances.
Considering solvers with arbitrary precision, we see \solver{c2d} by
Adnan Darwiche on the second place with 60 solved instances, and
\solver{DPMC} by Vu Phan, Jeffrey Dudek, and Moshe Vardi on the third
place with 43 solved instances.
Turning away to small precision loss, which allows developers to
participate if they were so far unable to integrate an arbitrary
precision library, we see that \solver{GPMC} by Kenji Hashimoto and
Shota Yap is the runner-up with 68 solved instances, closely followed
by \solver{D4} submitted by Pierre Marquis and Jean-Marie Lagniez who
solved 66 instances.
In 2023, the picture remains fairly similar. The trophy goes to
\solver{SharpSAT-TD} by Tuukka Korhonen and Matti Järvisalo with 75
solved instances, undefeated ahead of the other solvers with 8
instances more.
The second place, goes to GPMC by Kenji Hashimoto and Shota Yap, which
solves 62 instances.
\solver{ExactMC-Arjun} by Yong Lai, Zhenghang Xu, Minghao Yin, Kuldeep
S. Meel, and Roland H.C. Yap takes the third place with 59 solved
instances.
Interestingly, we can see quite a notable differences in number of
solved instances for the arbitrary precision solvers as the winning
solver is 13 instances ahead.
In the small precision loss setting, we see a slightly different ranking.
Still, \solver{SharpSAT-TD} is head, followed by \solver{D4}, and
again \solver{GPMC}.
\solver{D4} was submitted by Pierre Marquis and Jean-Marie Lagniez and
solved 67 instances, so 8 instances behind \solver{SharpSAT-TD} and 5
instances ahead of \solver{GPMC}.
Overall, we observe five very strong solvers with different underlying
techniques, namely, \solver{SharpSAT-TD}, \solver{D4}, \solver{GPMC},
\solver{ExactMC-Arjun}, and \solver{c2d}.

\begin{table}[t]
  \centering
  \begin{subtable}[t]{0.48\textwidth}
    \centering
    \resizebox{.95\textwidth}{!}{%
  \begin{tabular}{ccclHcc}
    \toprule
    0\% & 0.1\% & 0.8 & Submission           & Authors                                                              & solved \\
    \midrule
       &      & 1   & \solver{Arjun-GANAK-ApproxMC} & Mate Soos Kuldeep S. Meel                                            & 81     \\
    1  & 1    & 2   & \solver{D4}                   & Pierre MarquisJean-Marie Lagniez                                     & 71     \\
    2  & 2    & 3   & \solver{GPMC}                 & Kenji Hashimoto, Shota Yap  & 67     \\
    3  & 3    & 4   & \solver{Arjun-GANAK}          & Mate Soos Kuldeep S. Meel                                            & 64     \\
    4  & 4    & 5   & \solver{DPMC}                 & Vu Phan, Jeffrey Dudek, Moshe Vardi                                  & 36     \\
       & X    & X   & \solver{Alt-DPMC}             & Aditya Shrotri, Moshe Vardi & 12     \\
       & 5    & 6   & \solver{mtmc}                 & Ivor Spence                                                          & 4     \\
    \bottomrule
  \end{tabular}
  }%
  \caption{Track 3 (PMC) 2023}
  \end{subtable}
 \begin{subtable}[t]{0.48\textwidth}
   \centering
   \resizebox{.95\textwidth}{!}{%
    \begin{tabular}{lllHl}
      \toprule
      0\% & 0.8 & Submission & Authors                              & solved \\
      \midrule
       & 1 & \solver{GANAK (approx)} & Mate Soos and Kuldeep Meel & 83\\
      1 & 2 & \solver{GPMC}       & Kenji Hashimoto, Shota Yap           & 72     \\
      2 & 3&\solver{D4}         & Pierre Marquis, Jean-Marie Lagniez   & 71     \\
      3 & 4&\solver{GANAK (exact)}      & Mate Soos, Kuldeep Meel              & 56     \\
      4 & 5& \solver{DPMC}       & Vu Phan, Jeffrey Dudek, Moshe Vardi  & 26    \\
      \bottomrule
    \end{tabular}
    }%
    \caption{Track 3 (PMC) 2022}
  \end{subtable}
  \\[1em]
  \begin{subtable}[t]{0.48\textwidth}
    \centering
    \begin{tabular}{llHlHH}
      \toprule
      \# & Submission    & Authors                                   & solved & excl \\
      \midrule
      1  & \solver{GPMC}          & Kenji Hashimoto, Takaaki Isogai           & 70   & 0             \\
      2  & \solver{D4}            & Jean-Marie Lagniez, Pierre Marquis        & 57   & 0             \\
      3  & \solver{narasimha} & Sharma,  Lai, Xu,  Roy, Yap,  Soos,  Meel & 52$^\dagger$   & 22            \\
      4  & \solver{pc2bdd}        & Kenji Hashimoto, Takaaki Isogai           & 41   & 0             \\
      5  & \solver{ProCount}      & Vu PhanJeffrey Dudek Moshe Vardi          & 21   & 0             \\
      6  & \solver{c2d}           & Adnan Darwiche                            & 4    & 0             \\
      \bottomrule
    \end{tabular}
    \caption{Track 3 (PMC) 2021}
  \end{subtable}
  \caption{Results Track 3 (PMC). $^\dagger$: we excluded 22 instances,
    which were within 0.8 approximation factor but not within the expected
    margin for the 2021 iteration.
  }%
  \label{tab:summary:pmc}
\end{table}

\subsection{Track 3 (PMC)}\label{sec:results:pmc}
Table~\ref{tab:summary:pmc} states a detailed overview on the
standings and solvers in the years 2021, 2022, and 2023. We employ
rankings as previously given in Section~\ref{sec:ranking}.
In 2021, \solver{GPMC} by Kenji Hashimoto and Shota Yap solved in
total 70 instances and thereby scored the winning trophy. The
\solver{D4} solver by Pierre Marquis and Jean-Marie Lagniez ranks
second with 57 solved instances. The portfolio solver
\solver{narasimha}, which was submitted by Shubham Sharma,Yong Lai,
Zhenghang Xu, Subhajit Roy, Roland H.\ C.\ Yap, Mate Soos, and Kuldeep
S.\ Meel managed to solve 52 instances correctly. Note that
\solver{narasihma} solved 74 instances in total and would have even
been ahead of all other solvers, however, we had to exclude 22
instances from the ranking. These 22 instances were outside the
expected margin 1\% error, but clearly within the 0.8 approximation
factor that is usually assumed for approximate
counting~\cite{SoosMeel19,SharmaEtAl19a,ChakrabortyMeelVardi16a}.
Since our ranking in 2021 did not distinguish between approximate and
exact solving, we went into long discussions. We concluded to exclude
these instances, consider them as unsolved, to not disqualify the
solver, and revise the rules for the following iterations.
In 2022, Mate Soos and Kuldeep Meel submitted a new solver called
\solver{GANAK}, which managed to solve 83 instances in an approximate
configuration and 56 instances in an exact configuration.  Thereby
they win the approximate ranking and score third in the exact ranking.
Their approximate configuration was 11 instances ahead of the
previously best solver, which is a quite notable difference.
The \solver{GPMC} solver by Kenji Hashimoto and Shota Yap takes the
second place in the exact ranking with 72 instances.
Closely followed, we see the \solver{D4} solver by Pierre Marquis and
Jean-Marie Lagniez with 71 instances solved and hence just one
instance less.
In 2023, the ranking did not change much. The portfolio consisting of
\solver{GANAK}, \solver{ApproxMC}, and the preprocessor
\solver{Arjun}, which was submitted by Mate Soos and Kuldeep S. Meel
solved in total 81 instances. Thus, they are clear winner in the
approximate ranking. Comparing to exact solving techniques, we see 71
solved instances by \solver{D4} scoring second in approximate and
first in exact ranking, and 67 solved instances by \solver{GPMC}
resulting in the third and second place respectively.  Finally, the
combination of \solver{GANAK} and \solver{Arjun} solved 64 instances
and hence scored third in the exact ranking.
In summary, we see a clear development of extremely strong solvers for
projected model counting. Approximate solving can provide a notable
difference of solved instances when approximations are acceptable.
Authors compete hard and manage to improve every year and asks for the
scientific question whether improvement is primarily due to
engineering or new theoretical insights.

\begin{table}[t!]
  \centering
  \begin{subtable}[t]{0.48\textwidth}
    \centering
    \begin{tabular}{lllHl}
      \toprule
      0\% & 0.1\% & Submission & Authors & solved \\
      \midrule
      1 & 1 & \solver{GPMC} & Kenji Hashimoto, Shota Yap & 82 \\
          & 2 & \solver{D4} & Pierre Marquis, Jean-Marie Lagniez & 72 \\
      2 & 3 & \solver{DPMC} & Vu Phan, Jeffrey Dudek, Moshe Vardi & 39 \\
          & 4 & \solver{Alt-DPMC}$^\dagger$ & Aditya Shrotri, Moshe Vardi & 37\\
      \bottomrule
    \end{tabular}
    \caption{Track 4 (PWMC) 2023}
  \end{subtable}
      \begin{subtable}[t]{0.48\textwidth}
    \centering
    \begin{tabular}{lllHl}
      \toprule
      0\% & 0.1\% & Submission & Authors & solved\\
      \midrule
          & 1 & \solver{GPMC} & Kenji Hashimoto, Shota Yap           & 79 \\
      1 & 2 & \solver{DPMC} & Vu Phan, Jeffrey Dudek, Moshe Vardi & 35\\
      \bottomrule\\\\
    \end{tabular}
    \caption{Track 4 (PWMC) 2022}
  \end{subtable}
  \caption{Results Track 4 (PWMC). $^\dagger$ a parallel version of
    the solver solved 40 instances in total.}
  \label{tab:summary:pwmc}
\end{table}

\subsection{Track 4 (PWMC)}\label{sec:results:pwmc}
In 2022, we established a track that combines projected and weighted
model counting to open up for potential applications that require
weights projected to a certain subset of the variables.
Table~\ref{tab:summary:pwmc} provides an overview on the results in
the 2022 and 2023 iterations according to rankings as given in
Section~\ref{sec:ranking}.
In 2022, the \solver{GPMC} solver, which was submitted by Kenji
Hashimoto and Shota Yap, solved 79 instances scored second in the
arbitrary precision ranking and first in the small precision loss
ranking. Vu Phan, Jeffrey Dudek, and Moshe Vardi contributed
\solver{DPMC}, which outputted a solution to 35 instances in total.
In 2023, we received four submissions. \solver{GPMC} ranked first,
however, this time in both categories by solving 82 instances
successfully. \solver{D4}, submitted by Pierre Marquis and Jean-Marie
Lagniez solved 72 instances and scored second in the small precision
loss ranking. \solver{DPMC} solved 39 instances and scored second in
the arbitrary precision ranking and third in the small precision loss
ranking.
In summary, we received more contributions to the projected weighted
model counting track and were able to establish a new
problem. Furthermore, we were positively surprised that solvers
perform quite well.

\subsection{Summary}
Overall, we saw very unexpected improvements in the three years of the
competition. When \solver{SharpSAT-TD} entered the competition in
2021, everyone was surprised that a component caching-based solver
enhanced by tree decompositions would solve significantly more
instances than knowledge-compilation-based solvers or approximate
counters.
In addition, this development revived the underlying solver
\solver{SharpSAT}, which is based on
\solver{minisat}~\cite{EenMishchenkoSorensson07} and solves significantly less
instances than the two successful knowledge-compilers \solver{c2d} or
\solver{D4}.
Another big surprise was that approximate counting could not show its
performance over the exact counters within the limits of the
competition on Track~1 (MC). We suspect that the reason was our strict
requirements on approximation factor and size of the
instances. However, in Track~3 (PMC) approximate or probabilistic
techniques contributed to very successful solving portfolios. This
makes us believe that there is clearly a value for approximate
counting on larger instances and when we do not require exact
solutions.
For Track~3 (PMC), \solver{GPMC} showed very high quality and stable
performance throughout the years. Clearly, a very well developed
solver.
Unsurprisingly, the picture on Track~2 (WMC) is not much different to
Track~1 (MC). The reason is likely, that we generated instances
randomly from previous collections of instances from Track~1 (MC).
Due to a consistently missing number of submitted benchmark instances,
we suggest to abandon the track in the next iteration.
Track~4 (PWMC) received an increasing number of submissions, but also
shows a similar behavior as Track~3 (PMC) throughout the years. Hence,
we also suggest to abandon the track.

\paragraph{Solving Techniques}
We can divide the received submissions by their underlying solving
techniques. Interestingly, these vary quite a bit and are by far more
sophisticated than simply enumerating solutions.
In principle, we can classify solvers into four categories
(i)~search, %
(ii)~knowledge compilation, %
(iii)~dynamic programming, and %
(iv)~approximate counting. %
Solvers that belong to (i), (ii), or (iv) employ principles from SAT
solvers. %
Some submissions combine multiple techniques in a simple solving
portfolio.
To each category, we can find numerous research articles on basics,
implementations, and underlying principles.
Component-caching-based solving uses two main ideas: search paths that
cannot lead to a solutions are cut using clause learning and counts to
a subproblem are stored in caches, which can then be used to avoid
expensive recomputation~\cite{SangEtAl04,Thurley06a}. For practical
successful solving, the design of caches~\cite{LatourEtAl19,LatourBabakiFokkinga22} and balancing
search heuristics~\cite{KorhonenJarvisalo21a,KorhonenJarvisalo23a} are
important challenges.
Knowledge-compilation aims at transforming an input formula usually
given in CNF into an equivalent formula in another normal form on
which counting is efficient. Equivalence can be logical equivalence or
one can also think of an equivalence that just preserves certain
reasoning properties. There is extensive theory behind theoretical
properties of knowledge compilation~\cite{darwiche2002knowledge}.
Dynamic programming is based on splitting the search according to a
theoretically clearly defined structure. We are mostly well aware of
the theoretical benefits and limitations. However, underlying ideas
can also be combined with other solving techniques to guide solving
along specific decompositions~\cite{HecherThierWoltran20}.
Approximate solving aims at splitting the search-space uniformly and
then sampling solutions until a sufficiently probabilistic estimate of
the count can be outputted~\cite{ChakrabortyEtAl14a}.
While these techniques appear to be inherently different, they share
many similarities.  Recent
articles~\cite{MuiseEtAl12a,KieselEiter23a,EiterHecherKiesel21}
identify connections between structure from a theoretical perspective
and solving approaches.

\paragraph{Notable Solvers} %
During 2021--2023 iterations of the competition, we received highly
interesting new solvers. The following solvers drew our attention
throughout the years.
\solver{SharpSAT-TD}~\cite{KorhonenJarvisalo21a,KorhonenJarvisalo23a}
was submitted first in 2021 and turned out to be the best solver right
away solving 18 instances more than the previously best exact
solver. It solved also more instances than approximate model counters
on our instance sets for Track~1~(MC) and Track~2~(WMC).
Interestingly, authors of \solver{SharpSAT-TD} managed to keep the
performance up in all three iterations while only minor changes were
introduced.
The solver \solver{GPMC} stood out as one of the best solvers on
Track~3~(PMC) and also supports Track~4~(PWMC). It did not perform as
strong as the other solvers on the remaining tracks but showed quite
good performance. %
The best knowledge compiler turned out to be \solver{D4}, which allows
to easily output Decision-DNNFs that can then be used to count
multiple times. In the most recent iteration, \solver{D4} turned out
to be also the best exact solver on Track~3~(PMC).
\solver{ExactMC} turned out to be an interesting newcomer in the field
knowledge compilation~\cite{LaiMeelYap21a} being among the best
solvers.
After the first two competitions, it seemed that approximate counting
was of limited use over exact solvers on the selected competition
instances.
However, after the 2022 and 2023 iterations, we see that approximate
counting shows quite interesting performance on Track~3~(PMC) when
combined into a portfolio allowing to solve more instances than other
solvers.

\paragraph{Expected Model Counts, Errors, and Correctness}
To check whether solvers output the expected model counts, we
precompute counts using the three best counters from the previous
iteration.
For execution on StarExec, we consider only counts in log10 notation
and compare with a bash script at very low precision to the log10
estimates outputted by the solvers.
To validate whether the solvers are within the expected error ranges,
we download all outputs from StarExec, parse them separately for the
highest precision output, and run comparisons with an arbitrary
precision library.
After the competition, we additionally examine whether the solvers
agree on the model count and manually investigated for discrepancies
between solvers by inspecting the counts in log10-notation.
Obviously, a majority decision is insufficient for model counting and
holds the disadvantage that a these solvers are based on a specific
technique, which all have the same underlying bug.  Hence, a majority
check is only a necessary condition. A sufficient condition would
require to emit a correctness proof and verify the proof. Recent
advances on proof systems for model counting seem to be a promising
start~\cite{Capelli19,CapelliLagniezMarquis21a,FichteHecherRoland22a,BryantNawrockiAvigad23a,BeyersdorffHoffmannSpachmann23a}.
Unfortunately, only few solvers implement proof traces and these
traces can be quite large, which is a notable issue with our available
cluster resources.
Still, we discovered erroneous solvers and had to remove them
partially or entirely from rankings.
Most of the solvers worked fine on the provided instances, but we
discovered bugs after the competition on other instances.
Mate Soos also made us aware of issues when fuzzing the solvers.
For future editions, we suggest to introduce fuzzing and optional
proof traces and checking thereof, together with strict consequence of
disqualifying incorrect exact solvers instead of removing them from a
ranking.

\paragraph{Stability} Overall the stability of solvers improved
significantly within the three iterations of the competition. All
solvers manage to output the format correctly and only few solvers
produce segfaults. However, most solvers require manual flags when
running to select the type of instance and the start scripts are most
certainly not useful for the general public.
Moreover, some solvers do not interpret corner-cases in the format
specification correctly. For example, these solvers return differing
results when weights are given only for positive literals.
Housekeeping is still an
issue for some solvers, which output massive amounts of debug messages
or occupy notable amounts of disk space. This causes problems with
storage on StarExec and required us to split up jobs, manual download
results, and delete previous runs, which can easily affect stability
of evaluating results in the competition.
Finally, we would like to mention that we also improved stability on
our end, as we introduced format checking and correction tools.  We
discovered that some of the earlier bugs were caused by
ambiguities. Therefore, we checked all instances and also corrected
collected instances if needed.

\paragraph{Execution (StarExec)}
The StarExec system allowed us a notable amount of flexibility and
provided us with stable cluster resources on which developers could
easily debug potential errors.
Some extended requirements did not allow us to use leader boards or
outputs directly from the StarExec system.
Still, the system proved to be of major value and we were quite
discouraged to hear that the Logic solving communities will likely
loose StarExec as tool in 2025. Without joint efforts on computational
resources, future organizers will effectively be in the rain and
require to spend lots of hours to handle solver submissions and
debugging.
We hope that Markus Iser, who is currently leading an initiative, can
provide us with new insights and resources.
To our surprise, we obtained unstable outputs and runs when using
BenchExec~\cite{BeyerLoweWendler15,BeyerLoweWendler19} and therefore
employed runsolver~\cite{Roussel11} to limit resources on StarExec.

\paragraph{Open Source and Replication}
During the four years of the competition, \solver{c2d} was the only
non-open source solver.
Due to its absence in the latest iteration, the 2023 competition
contained only open source submissions.
To spare developers and researchers precious research time, we provide
a full system to replicate aspects of the competition using a tool
called
\href{https://github.com/tlyphed/copperbench}{\solver{copperbench}}~\cite{FichteEtAl24}.
\solver{copperbench} allows to generate executions for the widely used
high performance cluster (HPC) software Slurm~\cite{YooJetteGrondona03} and
provides a systematic approach to its evaluation.
We provide a data set containing basic configuration for this tool and
all solvers including the necessary wrappers.
A user can download \solver{copperbench}, a data set that contains a
basic setup for slurm-based clusters
\href{https://doi.org/10.5281/zenodo.10866053}{Zenodo:10866053}, and
instances from the
\href{https://zenodo.org/communities/modelcounting/}{Zenodo Model
Counting Community} and can generate execution jobs to run an
empirical evaluation with existing solvers on other clusters.
Note that the current setup is only tested on two cluster setups, so
we appreciate feedback.
StarExec exports are available at
\href{https://doi.org/10.5281/zenodo.10671987}{Zenodo:10671987}.
Note that this set may contain multiple StarExec runs, which was
inevitable as we had to run some solvers multiple times due to disk
space restrictions.

\section{Organization}
The composition of the program committee during the competitions
2021--2023 was as follows:

\begin{table}[h]
  \centering
  \begin{tabular}{llH}
    \textbf{Program Committee}
    &Johannes Fichte & Link\"oping University\\
    &Markus Hecher & Massachusetts Institute of Technology\\
    \textbf{Technical Advisor} & Daniel Le Berre\\
    \textbf{Judge} & Martin Gebser (2021-2023)\\
    \textbf{} & Mario Alviano (2022)\\
  \end{tabular}
\end{table}

\section{Conclusion and Future}\label{sec:concl}
Before we conclude, we would like to thank all participants. There
would have been no competition without their solvers, benchmark
instances, and new applications. Diverse benchmark collections and new
applications are crucial to test and challenge researchers for new
techniques.

\medskip
\noindent In this paper, we survey work on the model counting
competition of the 2021--2023 iterations. We introduce the novice
reader to the four main problems of interest, namely, model counting
(\MC), weighted model counting (\WMC), projected model counting
(\PMC), and projected weighted model counting (\PWMC). We list classic
and recent competition solvers, including supported problems,
competition years, scientific references, and download links
(Table~\ref{tab:mc-solvers}). We illustrate competition tracks and
explain the background on accuracy, precision, rankings, and computing
infrastructure.
We provide access to all collected instances, explain the selection
process, and briefly explain instance characteristics.
Then, we survey participants and the results of the three
competitions.
Throughout these years, we received extremely strong
contributions. New techniques challenged existing solvers. Algorithms
and their engineering improved notably. A new, fast \solver{Arjun}
preprocessor showed up, developers of knowledge compilers
(\solver{D4}) fought hard for improvements, tree decompositions spured
search-based solvers (\solver{SharpSAT-TD}), and multiple new solvers
appeared during these years (\solver{GPMC}, \solver{DPMC},
\solver{ExactMC} \solver{mtmc}, \solver{GANAK}, \solver{pc2dd},
\solver{ProCount}).
Moreover, we clearly see stability improvements and emerging interest
in certified counting from a theoretical
perspective~\cite{Capelli19,CapelliLagniezMarquis21a,FichteHecherRoland22a,BryantNawrockiAvigad23a,BeyersdorffHoffmannSpachmann23a}.
We conclude that the competition inspired both practical and
theoretical research and provides an annual event for researchers to
present new implementations.

\paragraph{Competition Instances}
The competition provides an interesting snapshot on the current state
of development of model counters, serves as important event to get
researchers together, and provide a joint platform for interested
potential users.
Hence, competition instances may serve as helpful baseline for future
comparisons.
However, given the sheer amount of instances and nature of
computational hardness of model counting, we face the difficulty to
select instances for the competition.  In turn, it poses a severe
danger to empirical evaluations by tracking progress and improvements
solely by considering these benchmark sets as sole truth.
But there might well be solvers that perform well on specific sets of
instances or application. These solvers may show high value while
performing bad on the overall competition benchmarks.
Therefore, we would like to encourage reviewers and authors to focus
on pros and cons of solvers instead of evaluating whether the solver
improved over one or all iterations of the model counting
competition. We believe in diversity of tools and
applications. Consequently, we also released the full set of
instances, which we collected or received.

\paragraph{Future Works}
There are many open topics for future research that originate in the
competition.
A replication imitative similar to SAT Heritage
project~\cite{AudemardPauleveSimon20a,Di-CosmoZacchiroli17a} could be
helpful to track progress of solvers over time and make them long term
accessible.
We expect that certified model counting will certainly be of huge
interest over the next years, especially under the light that counting
could prove helpful for applications in verification. In that line,
counting modulo satisfiability might be an interesting for
applications~\cite{SpallittaMasinaMorettin22}.
For the competition itself, we see the topic of effective
preprocessing~\cite{LagniezMarquis14a}, robust benchmark
selection~\cite{Smith-MilesMunoz23,KandanaarachchiSmith-Miles23},
stable rankings (number of solved instances or PAR-X ranking favoring
fast solvers as used in SAT competitions~\cite{Iser23}), new
applications, classification of instances, and
practical~\cite{PaivaMorenoSmith-Miles22} vs theoretical hardness of
solving as interesting targets for future research.
For applications, investigations into on solving portfolios or
algorithm configuration~\cite{Hoos12} might be interesting as well.

\paragraph{Next Iteration}
Preparations for the next iteration %
are already in process
and we hope for new interesting instances and many great solver
submissions.
We welcome anyone who is interested in the competition to send us an
email (\url{mailto:organizers@mccompetition.org}) for receiving
updates and joining the discussion. We look forward to the next
edition. Detailed information will be posted on the website at
\url{modelcounting.org}.

\section*{Acknowledgements}
\noindent The authors gratefully acknowledge the following computation
centers for resources:
\begin{itemize}
\item StarExec under the direction of Aaron Stump (Iowa), Geoff
  Sutcliffe (University of Miami), and Cesare Tinelli
  (Iowa). See: \url{https://www.starexec.org/starexec/public/about.jsp}
\item The GWK support for funding this project by providing computing
  time through the Center for Information Services and HPC (ZIH) at TU
  Dresden. See:
  \url{https://tu-dresden.de/zih/hochleistungsrechnen/nhr-center?set_language=en}
\item Computation resources provided by the National Academic
  Infrastructure for Supercomputing in Sweden (NAISS) at Link\"oping
  partially funded by the Swedish Research Council through grant
  agreement no. 2022-06725. See: \url{https://www.naiss.se/}
\end{itemize}

\noindent Furthermore, we would like to thank the following people for helping
with the organization and suggestion for future editions.
First, we would like to thank Daniel Le Berre for detailed, reliable,
and fast input and suggestions to make the competition a successful
community event.
Aron Stump made StarExec available to the model counting community and
allowed us to evaluate many solvers and instances in a uniform way. He
was available for bug fixing when we discovered problems that arose
during our evaluations.
Stefan Woltran, Toni Pisjak, Mathias Schl\"ogel, and Tobias Geibinger
provided cluster resources at TU Wien and Sarah Gaggl supported guest
access of resources at TU Dresden for the pre-evaluation phases and
for cleaning up instances.
Steffen H\"olldolber and Sarah Gaggl enabled us to use high
performance computing (HPC) resources on the Dresden data center for
the pre-computation phase and analyzing instance characteristics.
Finally, we would like to remember Toni Pisjak, Steffen H\"olldobler,
and Fahiem Bacchus who all passed away way by far too soon.
  Fahiem was one of the pioneers working on practical model counting.

\section*{Declaration of Competing Interest}
The authors of the report co-authored model counters called
\solver{DPDB}~\cite{FichteHecherThier22},
\solver{NestHDB}~\cite{FichteHecherMorak23}, and
\solver{gpusat}~\cite{FichteHecherRoland21}. These solvers never
participated in any competition and have not been evaluated along the
line of this work.
The authors have a joint publication together with Daniel Le
Berre~\cite{FichteBerreHecher23}.

\section*{Ethical Statement}
The work did neither involve humans or animals nor data on about them.

\section*{Declaration of generative AI in scientific writing}
The authors used generative AI (Grammarly) to  identify
potential language improvements.

\clearpage
\appendix
\section{2021+ Data Format  (DIMACS-like)}\label{sec:format}
For the 2020 competition, we suggested a uniform data format.
However, after the competition, we received multiple complaints about
issues in the format. In particular, that headers are not compatible
with SAT solvers, which requires reencoding and breaks downwards
compatibility.
In order to fix these issues, we reevaluated the format in 2021 and
suggested the format as listed below. We keeping the format very close
to the original DIMACS format, removing ambiguities, and keep SAT
compatibility.

\subsection{Input Format}
In the Model Counting Competition 2021, we use a DIMACS CNF-like input
format~\cite{TrickChvatalCook93a}. The format extends the format used
in SAT
competitions~\cite{JarvisaloBerreRoussel12a,sat_competition20}\footnote{See,
  for example,
  \url{http://www.satcompetition.org/2009/format-benchmarks2009.html}}
by introducing statements for weights and projections similar as in
Cachet~\cite{SangBeameKautz05a} or GANAK~\cite{SharmaEtAl19a}.

\medskip\noindent
The following description gives an idea on the expected input format:\vspace{-0.75em}
\begin{verbatim}
c c this is a comment and will be ignored
c c REPRODUCIBILITY LINE MANDATORY FOR SUBMITTED BENCHMARKS
c r originUrl/doi descUrl/doi [generatorUrl/doi]
c c HEADER AS IN DIMACS CNF
p cnf n m
c c OPTIONAL MODEL COUNTING HEADER
c t mc|wmc|pmc|pwmc
c c PROBLEM SPECIFIC LINES for wmc|pmc
c p ...
c p ...
c c CLAUSES AS IN DIMACS CNF
-1 -2 0
2 3 -4 0
c c this is a comment and will be ignored
4 5 0
4 6 0
\end{verbatim}

\noindent In more details (note that we print symbols in typewriter font):
\begin{itemize}
\item Line separator is the symbol \fsym{$\backslash{}$n}. Expressions 
  are separated by space.
\item A line starting with character \fsym{c r} is a comment. The
  solver may ignore it. The line aims for open data and
  reproducibility of the submitted instances expressing the origin of
  the benchmarks. The line will be added after competition when
  publishing the instances.
  \fsym{originUrl/doi} states where the instance can be downloaded
  (either as URL or DOI). \fsym{descUrl/doi} links to a description of
  the benchmarks. \fsym{generatorUrl/doi} provides an optional URL/DOI
  to where a problem generator can be found.
\item A line starting with character \fsym{c t} is a comment. The
  starting character will be followed by \fsym{mc}, \fsym{wmc},
  \fsym{pmc}, or \fsym{pwmc} indicating possible problem specific
  lines starting with \fsym{cs} in the file. If present, the line will
  occur prior to problem specific line. The solver may ignore it.
\item A line starting with character \fsym{c p} provides problem
  specific details for weighted or projected model counting. Lines may
  occur anywhere in the file. We assume that lines are consistent and
  provide no contradicting information. We provide more details on its
  meaning below.
\item Lines starting with character \fsym{c} \emph{followed by} a
  second \emph{character differing} from \fsym{p} and \fsym{s} are
  comments and can occur anywhere in the file. For convenience, we
  provide comments by lines starting with \fsym{c c}.
\item Variables are consecutively numbered from \fsym{1} to \fsym{n}.
\item The problem description is given by 
  a unique line of the form
  \fsym{p cnf NumVariables NumClauses} that we expect to be the first
  line (except comments). 
  More precisely, the line starts with character p (no other line may
  start with p), followed by the problem descriptor \fsym{cnf},
  followed by number \fsym{n} of variables followed by number \fsym{m}
  of clauses each symbol is separated by space each time.

\item The remaining lines indicate clauses consisting of decimal
  integers separated by space.  Lines are terminated by character
  \fsym{0}.  The Line \fsym{2 -1 3 0$\backslash{}$n} indicates the
  clause ``2 or not 1 or 3'' ($v_2 \vee \mneg v_1 \vee v_3$).  If more
  lines than announced are present, the solver shall return a parser
  error and terminate without solving the instance.
\item Empty lines or lines consisting of spaces/tabs only may occur
  and can be ignored.
\end{itemize}

\paragraph{Weighted Model Counting}
For weighted model counting, we introduce optional problem specific
lines:
\begin{verbatim}
c p weight  1 0.4 0 
c p weight -1 0.6 0 
c p ...
\end{verbatim}

\noindent In more details (note that we print symbols in typewriter font):
\begin{itemize}
\item %
  The weight function is given by lines of the form \fsym{c p weight
    $\ell$ $w_\ell$ 0} defining the weight~$w_\ell$ for literal~$\ell$,
  where $0 \leq w_\ell$.
  The weight will be given as floating point (e.g., \fsym{0.0003})
  with at most 9 significant digits after the decimal point, or in
  32-bit scientific floating point notation (e.g., \fsym{1.23e+4}), or as fraction (e.g.,
  \fsym{3/10}) consisting of two integers separated by the
  symbol~\fsym{/}.
\item We expect that the submission tests instances and output if it
  cannot be handled correctly. For example, if an instance specifies a
  function that has more than 9 significant digits (e.g.,
  \fsym{0.00000000009}), larger floating point values, or large
  fractional values. In such a case, the solver should output a
  parsing error as specified in the return code section.
\item We provide only instances where $0 \leq w_\ell \leq 1$ and
  $w_{\mneg \ell} + w_{\ell} = 1$ or $w_{\mneg \ell} = w_{\ell} = 1$.
\item If a weight~$w_\ell$ is defined for a literal~$\ell$ but
  $\mneg \ell$ is not given or $\mneg x$, we assume
  $w_{\mneg \ell} = 1 -w_\ell$ (assume that $\mneg \mneg a = a$).
In other words:
\begin{verbatim}
c p weight  1 0.4 0 
\end{verbatim}
or
\begin{verbatim}
c p weight -1 0.6 0 
\end{verbatim}
are the same as
\begin{verbatim}
c p weight  1 0.4 0 
c p weight -1 0.6 0 
\end{verbatim}
\item If the solver can handle weights $w_\ell > 1$ or
  $w_{\ell} + w_{\mneg \ell} > 1$ unless $w_{\ell}=w_{\mneg \ell}=1$, the
  solver must output a warning. Otherwise, the solver must output an
  error on those instances.
\item For compatibility, with \#SAT we say that if for a variable~$x$,
  there is neither a weight for $x$ nor
  $\mneg x$ given, it is considered 1.\\
  \emph{Note: this differs from the format used in Cachet.}
\end{itemize}

\paragraph{Projected Model Counting}
For projected model counting, we introduce optional problem specific
lines:
\begin{verbatim}
c c INDICATES VARIABLES THAT SHOULD BE USED
c p show varid1 varid2 ... 0 
....
c c MORE MIGHT BE GIVEN LATER
c p show varid7 varid23 ... 0
\end{verbatim}

\noindent In more details (note that we print symbols in typewriter font):
\begin{itemize}
\item %
  Projection variables will be given by lines starting with \fsym{c p
    show} followed by the identifier of the variables.
  Lines describing to add variables to a projection set may occur
  anywhere in the files and will be terminated by symbol \fsym{0}.
\end{itemize}

\noindent
Note: if all variables are stated using \fsym{show}, we consider model
counting. if no variables are stated the problem is simply to decide
satisfiability.

\subsection{Output Format}\label{sec:format:output}
We expect that the solver outputs result to \fsym{stdout} in the
following format.

\begin{verbatim}
c o This output describes a result of a run from 
c o a [weighted|projected] model counter.
c o 
c o The following line keeps backwards compatibility 
c o with SAT solvers and avoids underflows if result is 0.
c o MANDATORY
s SATISFIABLE|UNSATISFIABLE|UNKNOWN
c o The following solution line is optional.
c o It allows a user to double check whether a solver 
c o that provides multiple options was called correct.
c o
c o MANDATORY
c s type [mc|wmc|pmc|pwmc]
c o The solver has to output an estimate in scientific notation
c o on the solution size, even if it was an exact solver.
c o MANDATORY
c s log10-estimate VALUE
c o The solver outputs the solution in its highest precision.
c o MANDATORY
c s SOLVERTYPE PRECISION NOTATION VALUE
c o OPTIONAL
c o InternalValueForOpt=X
c o Internal=X
\end{verbatim}

\noindent In more details:
\begin{itemize}
\item While the solver may use a line staring with \fsym{c} for
  comments in the output, we suggest to use lines starting with
  \fsym{c o} to indicate a comment.
\item The solver has to announce whether the instance is satisfiable
  or unsatisfiable by a line starting with \fsym{s} followed by
  \fsym{SATISFIABLE} or \fsym{UNSATISFIABLE}. The solver may output
  \fsym{UNKNOWN}, but is not allowed to output any other value than
  these three if a line starting with \fsym{s} is present.
\item The solver has to announce an estimate on the solution by a line
  starting with \fsym{c s log10-estimate VALUE} where \fsym{VALUE} is
  a string representing the result (model count/weighted model
  count/projected model count) in $\log_{10}$-Notation (see
  Section~\ref{sec:prelimns:precs}) of double %
  precision,~i.e., 15 significant digits.
\item The output is has to be announced by a line starting with
  \fsym{c s} followed by strings indicating the solver type, the
  precision, the notation, and the result. The solver developer may
  use for \fsym{SOLVERTYPE} the following strings: \fsym{approx} and
  \fsym{exact}.
  In place of \fsym{PRECISION}, the solver developer has to specify
  which the internal precision the solver used, allowed values are
  \fsym{arb}, \fsym{single}, \fsym{double}, \fsym{quadruple}, or other
  values according to IEEE754 Standard.
  For \fsym{NOTATION}, the developer may chose \fsym{log10},
  \fsym{float}, \fsym{prec-sci}, or \fsym{int}.
  Then, in place of \fsym{VALUE}, the solver has to output its
  computed result.

  \emph{Note: If the solver developer announces as output a higher
    precision than the actual solver theoretically allows, we reserve
    the right to disqualify all submissions by the team.}
\item The solver may output a result in \emph{$\log_{10}$ notation} which
  is similar to the format used in the probabilistic inference
  competitions UAI~\cite{GogateEtAl16a}.
  For computing the relative accuracy of the solution of a solver, we
  refer to Section~\ref{sec:measure}.
\item If the solver consists of a run script, which calls a
  pre-processor, consists of multiple phases, or consists of a solving
  portfolio, we expect the developer to output by a comment line which
  tool was started, what parameters it used, and when the tool
  ended. Preferably as follows:\\
  \fsym{c o CALLS(1) ./preproc -parameters}\\
  \fsym{c o stat CALL1 STARTED RFC3339-TIMESTAMP}\\
  $\ldots$\\
  \fsym{c o stat CALL1 FINISHED RFC3339-TIMESTAMP}\\
  \fsym{c o CALLS(2) ./postproc -parameters}\\
  $\ldots$
\item We suggest that the solver outputs internal statistics by lines
  starting with \fsym{c o DESCRIPTION=VALUE} or \fsym{c DESCRIPTION :
    VALUE}.
\end{itemize}

\paragraph{Projected Weighted Model Counting}
The format for projected weighted model counting instances combines
the format of the weighted model counting instances and projected
model counting instances. The model counting header indicates
``\texttt{c t pwmc}''. The output is expected to contain a line
``\texttt{c s type pwmc}''.

\subsection{Examples}
The following sections provide a few brief examples for each track
with expected input and output.

\paragraph{Model Counting}
\begin{example}
  \noindent 
  The following text describes the CNF formula (set of clauses)
  \[\{\{\mneg x_1, \mneg x_2\}, \{x_2, x_3, \mneg x_4\}, \{x_4, x_5\},
    \{x_4, x_6\}\}.\]
\begin{verbatim}
c c This file describes a DIMACS-line CNF in MC 2021 format 
c c The instance has 6 variables and 4 clauses.
p cnf 6 4
c t mc
-1 -2    0
 2  3 -4 0
c c This line is a comment and can be ignored.
4 5 0
c The line contains a comment and can be ignored as well.
4 6 0
\end{verbatim}

\noindent A solution is given as follows, but can also be modified according 
to the technique of the solver:
\begin{verbatim}
c o This file describes a solution to a model counting instance.
s SATISFIABLE
c s type mc
c o The solver log10-estimates a solution of 22.
c s log10-estimate 1.342422680822206
c o Arbitrary precision result is 22.
c s exact arb int 22
\end{verbatim}
  
\end{example}

\paragraph{Weighted Model Counting}

\begin{example}
  \noindent 
  The following text describes the CNF formula (set of clauses)
  \[\{\mneg x_1, \mneg x_2\}, \{x_2, x_3, \mneg x_4\}, \{x_4, x_5\},
    \{x_4, x_6\}\}\] with weight
  function~$\{x_1 \mapsto 0.4, \mneg x_1 \mapsto 0.6, x_2 \mapsto 0.5,
  \mneg x_2 \mapsto 0.5, x_3 \mapsto 0.4, \mneg x_3 \mapsto 0.6, x_4
  \mapsto 0.3, \mneg x_4 \mapsto 0.7, x_5 \mapsto 0.5, \mneg x_5 \mapsto
  0.5, x_6 \mapsto 0.7, \mneg x_6 \mapsto 0.3\}$.
  \newline

\begin{verbatim}
c c This file describes a weighted CNF in MC 2021 format 
c c with 6 variables and 4 clauses 
p cnf 6 4
c t wmc
c c Weights are given as follows, spaces may be added 
c c to improve readability.
c p weight  1 0.4 0
c p weight  2 0.5 0
c p weight  3 0.4 0
c p weight  4 0.3 0
c p weight  5 0.5 0
c p weight  6 0.7 0
-1 -2 0
 2  3 -4 0
c this is a comment and will be ignored
 4  5 0
 4  6 0
c same
\end{verbatim}
  
\noindent The solution should be given in the following format (modified according to the technique of the solver):
\begin{verbatim}
c o This file describes a solution to a weighted 
c o model counting instance.
s SATISFIABILE
c s type wmc
c o This file describes that the weighted model count is 0.345
c o 
c s log10-estimate -0.460924
c s exact double float 0.346
\end{verbatim}
\end{example}  

\begin{example}[Optional]
  \noindent 
  The following text describes the CNF formula (set of clauses)
  \[\{\mneg x_1, x_2\}, \{x_3, \mneg x_2\}, \{x_2, x_1\},
    \{x_3, x_2\}\}\] with weight
  function~$\{x_1 \mapsto 0.1, \mneg x_1 \mapsto 0.1, x_2 \mapsto 0.1,
  \mneg x_2 \mapsto 0.9,\allowbreak x_3 \mapsto 0.0235,\allowbreak \mneg x_3 \mapsto 0.0125\}$
  including a problem description line and two comments.
  \newline

\begin{verbatim}
c c This file describes a weighted CNF in MC 2021 format 
c c with 3 variables and 4 clauses 
p cnf 3 4
c t wmc
-1  2 0
 3 -2 0
 2  1 0
 3  2 0
c p weight  1 0.1
c p weight -1 0.1
c p weight  2 0.1
c p weight  3 0.0235
c p weight -3 0.0125
\end{verbatim}

\noindent The solution should be given in the following format:  
\begin{verbatim}
c o WARNING
c o  L9:Sum of positive and negative literal is not equal to 1.
c o WARNING
c o  L12:Sum of positive and negative literal is not equal 1.
c o This file describes a solution to a weighted 
c o model counting instance.
s SATISFIABILE
c s type wmc
c o This file describes that the weighted model count is
c o 0.0004700000000000000532907051822
c s type wmc 
c s log10-estimate -3.327902142064282
c s exact arb log10 -3.3279021420642824863435269891
\end{verbatim}
\end{example}

\paragraph{Projected Model Counting}

\begin{example}
  \noindent 
  The following text describes the CNF formula (set of clauses)
  \[\{\mneg x_1, \mneg x_2\}, \{x_2, x_3, \mneg x_4\}, \{x_4, x_5\},
    \{x_4, x_6\}\}\] with projection set~~$\{x_1, x_2\}$ including a
  problem description line and two comments.
  \newline

\begin{verbatim}
c c This file describes a projected CNF in MC 2021 format
c c with 6 variables and 4 clauses and 2 projected variables 
p cnf 6 4 2
c t pmc
c p show 1 2
-1 -2 0
 2  3 -4 0
cc this is a comment and will be ignored
 4  5 0
 4  6 0
\end{verbatim}

\noindent A solution can be given in the following format:
\begin{verbatim}
c o This file describes that the projected model count is 3
s SATISFIABILE
c s log10-estimate 0.47712125471966
c s type pmc
c s exact arb int 3
\end{verbatim}
\end{example}

\subsection{Format Checking and Instance Generation}\label{sec:appendix:generator}
\paragraph{Checking and Converting}
We refer to the github repository at
\url{https://github.com/daajoe/mc_format_tools}.
The repository contains scripts for simple format checking
(\texttt{check\_mc2021\_format.py}), merging broken lines, correcting
typelines, construct valid instances from outputs of preprocessors,
converting instances from the 2020 format as well as the format used
by Daniel Fremont.
\paragraph{Instance Generator}
To obtain additional instances, we constructed projected model
counting instances by randomly choosing projection variables from
selected instances from Track 1 (MC). For Track 2 (WMC), we followed
two approaches. We selected instances based on sampling from weighted
model counting instances and model counting instances. For the model
counting instances, we first tried to construct a counting graph using
the \solver{D4} solver, which we then attributed randomly by weights
aiming for sufficiently large but not too small weighted counts. If we
could not compute a counting graph, we generated weights entirely
random.

\section{Platform Specifications}\label{sec:appendix:platform}

\begin{enumerate}
\item \system{StarExec} is a HPC system for computational
  competitions~\cite{StumpSutcliffeTinelli14}, see
  \href{https://www.starexec.org/starexec/public/machine-specs.txt}{\nolinkurl{starexec.org/starexec/public/machine-specs.txt}}.
  Our evaluation uses the 192 ``original'' nodes with the following specifications: \\
  \fsym{CPU:~~~~~~~~~~1x~Intel(R) Xeon(R) CPU E5-2609 0 @2.40GHz} \\
  \fsym{\mbox{~}~~~~~~~~~~~~~(4 cores, 4 memory channels)}\\
  \fsym{Cache Size:~~~10240  KB}\\
  \fsym{Main Memory:~~256~GB}\\ 
  \fsym{OS:~~~~~~~~~~~CentOS Linux release 7.7.1908 (Core)}\\ 
  \fsym{Kernel:~~~~~~~3.10.0-1062.4.3.el7.x86\_64}\\
  \fsym{glibc:~~~~~~~~glibc-2.17-292.el7.x86\_64, glibc-2.17-292.el7.i686}\\
  \fsym{gcc:~~~~~~~~~~gcc-4.8.5-39.el7.x86\_64}

\item \system{Taurus} is part of a TOP500 HPC system for scientific
  purposes~\cite{taurus}, see
  \href{https://doc.zih.tu-dresden.de/}{\nolinkurl{doc.zih.tu-dresden.de}}.
  We employed the Haswell nodes with the following specifications:\\
  \fsym{CPU:~~~~~~~~~~~2x~Intel Xeon E5-2680 v3 @2.50GHz}\\
  \fsym{\mbox{~}~~~~~~~~~~~~~~(12 cores, 4 memory channels)}\\
  \fsym{Cache Size:~~~~30720 KB}\\
  \fsym{Main Memory:~~~64~GB}\\
  \fsym{OS:~~~~~~~~~~~~Red Hat Enterprise Linux Server 7.9 (Maipo)}\\
  \fsym{glibc:~~~~~~~~~glibc-2.17-317.el7.x86\_64, glibc-2.17-317.el7.i686}\\
  \fsym{gcc:~~~~~~~~~~~gcc-4.8.5-44.el7.x86\_64}

\item \system{Cobra} is a small institute cluster for computational
  experiments consisting of 12 nodes in the following specification:\\
  \fsym{CPU:~~~~~~~~~~~2x~Intel(R) Xeon(R) CPU E5-2650 v4 @2.20GHz}\\
  \fsym{\mbox{~}~~~~~~~~~~~~~~(12 cores, 4 memory channels)}\\
  \fsym{Cache Size:~~~~30720 KB}\\
  \fsym{Main Memory:~~~256~GB}\\
  \fsym{OS:~~~~~~~~~~~~Ubuntu 16.04.1 LTS }\\
  \fsym{Kernel:~~~~~~~~Linux 4.4.0-184-generic}\\
  \fsym{glibc:~~~~~~~~~Ubuntu GLIBC 2.23-0ubuntu11.2 2.23}\\
  \fsym{gcc:~~~~~~~~~~~Ubuntu 5.4.0-6ubuntu1~16.04.12 5.4.0 20160609}
\item \system{Desktop} is a compact desktop system:\\
  \fsym{CPU:~~~~~~~~~~~1x Intel(R) Core(TM) i7-10710U CPU @1.10GHz}\\
 \fsym{\mbox{~}~~~~~~~~~~~~~~(6 cores, 2 memory channels)}\\
  \fsym{Cache size:~~~~12288 KB}\\
  \fsym{Main Memory:~~~64GB}\\
  \fsym{OS:~~~~~~~~~~~~Ubuntu 20.04.6 LTS}\\
  \fsym{Kernel:~~~~~~~~5.4.0-148-generic}\\
  \fsym{glibc:~~~~~~~~~GLIBC 2.31-0ubuntu9.9}\\
  \fsym{gcc:~~~~~~~~~~~Ubuntu 9.4.0-1ubuntu1~20.04.1 9.4.0}
\end{enumerate}

\section{Submissions and Authors}\label{appendix:submissions}
\begin{table}[h!]
  \centering
  \begin{tabular}{HHHp{12em}p{22em}H}
    \toprule
    0\% & 0.1\% & 0.8 & Submission           & Authors                                                                 & solved \\
    \midrule
        & 6     & 7   & \solver{Alt-DPMC}             & Aditya Shrotr,  Moshe Vardi                                             & 50     \\
        &       & 2   & \solver{Arjun-GANAK-ApproxMC} & Mate Soos, Kuldeep S. Meel                                              & 79     \\
        & 4     & 5   & \solver{Arjun-GANAK}          & Mate  Soos,  Kuldeep S. Meel                                             & 71     \\
    2   & 2     & 3   & \solver{D4}                   & Pierre Marquis, Jean-Marie Lagniez                                      & 75     \\
    4   & 7     & 8   & \solver{DPMC}                 & Vu Phan, Jeffrey Dudek, Moshe Vardi                                     & 43     \\
        & 3     & 4   & \solver{ExactMC-Arjun}        & Yong Lai, Zhenghang Xu, Minghao Yin,  Kuldeep S. Meel,  Roland H.C. Yap & 74     \\
    3   & 5     & 6   & \solver{GPMC}                 & Kenji Hashimoto,  Shota Yap                                             & 64     \\
    1   & 1     & 1   & \solver{SharpSAT-TD}          & Tuukka Korhonen, Matti Järvisalo                                        & 82     \\
        & 8     & 9   & \solver{mtmc}                 & Ivor Spence                                                             & 34     \\
    \bottomrule
  \end{tabular}
  \caption{Authors of the Submissions. 2023 Iteration}
  \label{tab:mcauthors}
\end{table}

\begin{table}[h!]
  \begin{tabular}{HHHp{12em}p{22em}H}
    \toprule
    0\% & 0.1\% & 0.8 & Submission                 & Authors                                                  & solved \\
    \midrule
    8   & 9     &     & \solver{c2d}                        & Adnan Darwiche                                           & 50     \\
    4   & 4     &     & \solver{D4}                         & Pierre Marquis, Jean-Marie Lagniez                       & 76     \\
    7   & 7     &     & \solver{DPMC}                       & Vu Phan, Jeffrey Dudek, Moshe Vardi                      & 61     \\
    2   & 2     &     & \solver{ExactMC}                    & Yong Lai, Kuldeep S. Meel. Roland H.C. Yap, Zhenghang Xu & 77     \\
        &       & 1   & \solver{SharpSAT-TD-Arjun\-ApproxMC} & Mate Soos, Kuldeep S. Meel                               & 74     \\
     1  & 1     &     & \solver{GANAK (approx)} & Mate Soos and Kuldeep Meel & 83\\
     1  & 1     &     & \solver{GANAK (exact)} & Mate Soos and Kuldeep Meel & 83\\
    5   & 5     &     & \solver{GPMC}                       & Kenji Hashimoto, Shota Yap                               & 69     \\
    6   & 6     &     & \solver{mtmc}                       & Ivor Spence                                              & 66     \\
    3    &  3     &     & \solver{SharpSAT-TD}                & Tuukka Korhonen, Matti Järvisalo                         & 77     \\
    1   & 1     &     & \solver{SharpSAT-TD-Arjun}          & Mate Soos Kuldeep S. Meel                                & 79     \\
        & 8     &     & \solver{TwG}                        & Sylvester Swats                                          & 53     \\
    \bottomrule
  \end{tabular}
  \caption{Authors of the Submissions. 2022 Iteration}
\end{table}

\begin{table}[t!]
  \begin{tabular}{Hp{12em}p{22em}HHH}
    \toprule
    \# & Submission    & Authors                                      & solved & excl \\
    \midrule
    8  & \solver{bob}           & Daniel Pehoushek                             & 11     & 0    \\
    7  & \solver{c2d}           & Adnan Darwiche                               & 29     & 0    \\
    3  & \solver{D4}            & Jean-Marie Lagniez,  Pierre Marquis          & 51     & 0    \\
       & \solver{DPMC}          & Vu Phan, Jeffrey Dudek, Moshe Vardi          & 34     & 0    \\
    4  & \solver{GPMC}          & Kenji Hashimoto, Takaaki Isogai              & 38     & 0    \\
    5  & \solver{MC2021\_swats} & Sylwester Swat                               & 34     & 0    \\
    2  & \solver{narasimha} & Shubham Sharma Yong Lai Zhenghang Xu Subhajit Roy
Roland H.C. Yap Mate Soos Kuldeep S. Meel & 61$^\dagger$     & 1    \\
      5  & \solver{ProCount}      & Vu Phan, Jeffrey Dudek, Moshe Vardi          & 21   & 0             \\
    1  & \solver{SharpSAT-TD}   & Tuukka Korhonen,  Matti Järvisalo            & 78     & 0    \\
    9  & \solver{SUMC2}         & Ivor Spence                                  & 7      & 0    \\
    \bottomrule
  \end{tabular}
  \caption{Authors of the Submissions. 2021 Iteration}

\end{table}

\clearpage


\end{document}